\documentclass[10pt, conference]{IEEEtran}
\usepackage[para,online,flushleft]{threeparttable}
\usepackage{graphicx}
\usepackage{balance}  
\usepackage{url}
\usepackage{booktabs}
\usepackage{etoolbox}
\usepackage{tabularx}
\makeatletter
\patchcmd{\maketitle}{\@copyrightspace}{}{}{}
\makeatother

\let\labelindent\relax
\usepackage{hyperref}       
\usepackage{amsmath}
\usepackage{amssymb}
\usepackage{leftidx}
\usepackage{bm}
\usepackage{comment}
\usepackage{color}
\usepackage{enumitem}
\usepackage{amsthm}
\usepackage[numbers]{natbib}
\usepackage[ruled, vlined, linesnumbered]{algorithm2e}
\usepackage{multicol}
\usepackage{graphicx}
\usepackage{capt-of}
\usepackage{etoolbox}
\usepackage{subcaption}
\usepackage{wrapfig}
\makeatletter
\patchcmd{\@makecaption}
  {\scshape}
  {}
  {}
  {}
\makeatother

\theoremstyle{remark}

\newtheoremstyle{problemstyle}  
        {3pt}                                               
        {3pt}                                               
        {\normalfont\itshape}                               
        {}                                                  
        {\bfseries}                 
        {\normalfont\bfseries:}         
        {.5em}                                          
        {}                                                  
\theoremstyle{problemstyle}

\makeatletter
\newcommand*{\transpose}{%
  {\mathpalette\@transpose{}}%
}
\newcommand*{\@transpose}[2]{%
  \raisebox{\depth}{$\m@th#1\intercal$}%
}
\makeatother

\usepackage{xcolor}
\usepackage{amsbsy}
\usepackage{subcaption}
\usepackage{enumitem}
\usepackage{amsmath,bm}
\usepackage{amsmath}
\usepackage{soul}
\usepackage{multicol}
\usepackage{multirow}

\usepackage{thmtools,thm-restate}
\newtheorem{case}{Case}

\newcommand{\x}{\textbf{x}}

\newcommand{\w}{\textbf{w}}

\DeclareMathOperator*{\argmin}{arg\,min}
\newcommand{\feng}{\textcolor{blue}}

\begin{document}
\IEEEoverridecommandlockouts
\title{How Out-of-Distribution Data Hurts Semi-Supervised Learning} \vspace{0\baselineskip}
\author{\IEEEauthorblockN{Xujiang Zhao* \thanks{* Equal Contribution. This work is done when Xujiang Zhao was with The University of Texas at Dallas.}}
\IEEEauthorblockA{NEC Laboratories America\\
xuzhao@nec-labs.edu}
\and
\IEEEauthorblockN{Killamsetty Krishnateja*, Rishabh Iyer, Feng Chen}
\IEEEauthorblockA{The University of Texas at Dallas\\
\{krishnateja.killamsetty,rishabh.iyer,feng.chen\}@utdallas.edu}
\vspace{-10mm}
}
\vspace{-10mm}

\maketitle
 \begin{abstract}
 Recent semi-supervised learning algorithms have demonstrated greater success with higher overall performance due to better-unlabeled data representations. Nonetheless, recent research suggests that the performance of the SSL algorithm can be degraded when the unlabeled set contains out-of-distribution examples (OODs). This work addresses the following question: \textit{How do out-of-distribution (OOD) data adversely affect semi-supervised learning algorithms?} To answer this question, we investigate the critical causes of OOD's negative effect on SSL algorithms. In particular, we found that 1) certain kinds of OOD data instances that are close to the decision boundary have a more significant impact on performance than those that are further away, and 2) Batch Normalization (BN), a popular module, may degrade rather than improve performance when the unlabeled set contains OODs. In this context, we developed a unified weighted robust SSL framework that can be easily extended to many existing SSL algorithms and improve their robustness against OODs. More specifically, we developed an efficient bi-level optimization algorithm that could accommodate high-order approximations of the objective and scale to multiple inner optimization steps to learn a massive number of weight parameters while outperforming existing low-order approximations of bi-level optimization. Further, we conduct a theoretical study of the impact of faraway OODs in the BN step and propose a weighted batch normalization (WBN) procedure for improved performance. Finally, we discuss the connection between our approach and low-order approximation techniques. Our experiments on synthetic and real-world datasets demonstrate that our proposed approach significantly enhances the robustness of four representative SSL algorithms against OODs compared to four state-of-the-art robust SSL strategies.
\end{abstract}

\section{Introduction}
Deep learning approaches have been shown to be successful on several supervised learning tasks, such as computer vision~\cite{dong2022neural,wang2021clear}, natural language processing~\cite{xu2021boosting}, and speech recognition~\cite{zhao2022seed}. However, these deep learning models are data-hungry and often require massive amounts of labeled examples to obtain good performance. Obtaining high-quality labeled examples can be very time-consuming and expensive, particularly where specialized skills are required in labeling (for example, in cancer detection on X-ray or CT-scan images). As a result, semi-supervised learning (SSL) has emerged as a very promising direction, where the learning algorithms try to effectively utilize the large unlabeled set (in conjunction) with a relatively small labeled set. Several recent SSL algorithms have been proposed for deep learning and have shown great promise empirically. These include Entropy Minimization ~\cite{grandvalet2005semi}, pseudo-label based methods ~\cite{lee2013pseudo, arazo2019pseudo,berthelot2019mixmatch}
and consistency based methods ~\cite{sajjadi2016regularization,laine2016temporal,tarvainen2017mean,miyato2018virtual} to name a few.

\begin{figure}[!t]
 \centering
 \hspace*{\fill}%
  \begin{subfigure}{0.35\textwidth}     
    \centering
    \includegraphics[width=\textwidth]{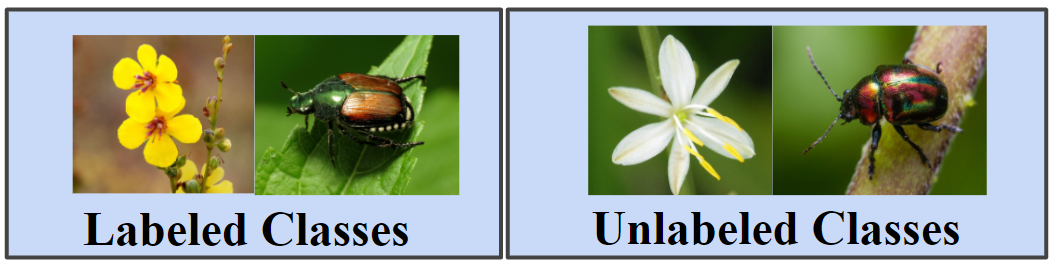}%
    \caption{Traditional semi-supervised learning }
    \label{fig:demo1}
  \end{subfigure}
  \hspace*{\fill}


  \hspace*{\fill}%
  \begin{subfigure}{0.47\textwidth}        
    \centering
    \includegraphics[width=\textwidth]{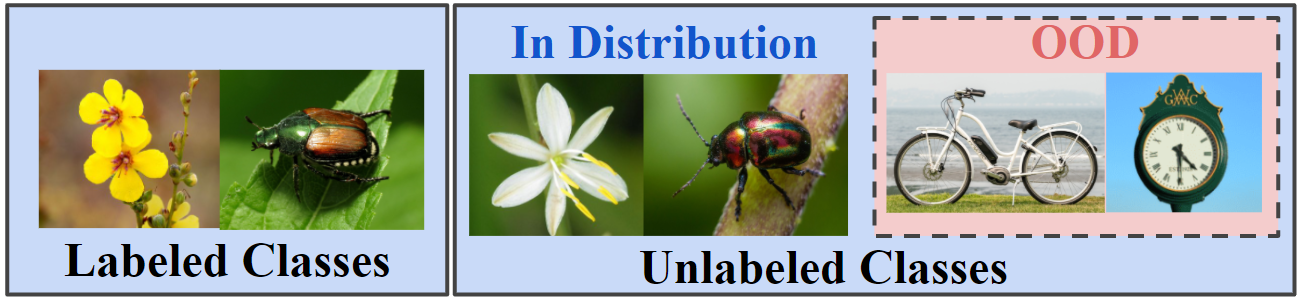}%
    \caption{Semi-supervised learning with OOD data}
    \label{fig:demo2}
  \end{subfigure}%
  \vspace{-2mm}
  \caption{(a) Traditional SSL. (b) SSL with OODs.}
  \label{fig:example}
  \vspace{-7mm}
\end{figure}

Despite the success of these SSL methods, they are designed with the assumption that labeled and unlabeled data have the same distribution. Fig~\ref{fig:example} (a) shows an example of this. However, this assumption may not hold in many real-world applications, such as web classification and medical diagnosis, where some unlabeled examples are from novel classes unseen in the labeled data. For example, Fig~\ref{fig:example} (b) illustrates an image classification scenario with out-of-distribution data, where the unlabeled dataset contains two novel classes (bicycle and clock) compared to the in-distribution classes (flower and beetle) in the labeled dataset. When the unlabeled set contains OOD examples (OODs), deep SSL performance can degrade substantially and is sometimes even worse than simple supervised learning (SL) approaches~\cite{oliver2018realistic}. Moreover, it is unreasonable to expect a human to go through and clean a large and massive unlabeled set in such cases.

A typical approach to robust SSL against OODs is to assign a weight to each unlabeled example based on some criteria and minimize a weighted training or validation loss. In an ideal weighting scheme, positive weights should be assigned only to ID samples while zero weights should be assigned to OOD samples. ~\citet{yan2016robust} applied a set of weak annotators to approximate the ground-truth labels as pseudo-labels to learn a robust SSL model.~\cite{chen2019distributionally} proposed a distributionally robust model that estimates a parametric weight function based on both the discrepancy and the consistency between the labeled data and the unlabeled data.
\cite{chen2020semi} (UASD) proposed to weigh the unlabeled examples based on an estimation of predictive uncertainty for each unlabeled example. The goal of UASD is to discard potentially irrelevant samples having low confidence scores and estimate the parameters by optimizing a regularized training loss.

A state-of-the-art method~\cite{guo2020safe}, called DS3L, considers a shallow neural network to predict the weights of unlabeled examples and estimate the parameters of the neural network based on a clean labeled set via bi-level optimization. 
It is common to obtain a dataset composed of two parts, including a relatively small but accurately labeled set and a large but coarsely labeled set from inexpensive crowd-sourcing services or other noisy sources. 

There are three \textbf{main limitations} of DS3L and other methods as reviewed above. First, it lacks a study of potential causes about the impact of OODs on SSL, and as a result, the interpretation of robust SSL methods becomes difficult. Second, existing robust SSL methods did not consider the negative impact of OODs on the utilization of BN in neural networks, and as a result, their robustness against OODs degrades significantly when a neural network includes BN layers. The utilization of BNs for deep SSL has an implicit assumption that the labeled and unlabeled examples follow a single or similar distributions, which is problematic when the unlabeled examples include OODs~\cite{ioffe2015batch}.
Third, the bi-level learning algorithm developed in DS3L relies on low-order approximations of the objective in the inner loop due to vanishing gradients or memory constraints. As a result of not using the high-order loss information, the learning performance of DS3L could be significantly degraded in some applications, as demonstrated in our experiments. Our main technical contributions over existing methods are summarized as follows:   

\noindent \textbf{The effect of OOD data points.} The first critical contribution of our work (Sec.~\ref{sec:Impact_OOD}) is to analyze what kind of OOD unlabeled data points affect the performance of SSL algorithms. In particular, we observe that OOD samples lying close to the decision boundary have more influence on SSL performance than those far from the boundary. Furthermore, we observe that the OOD instances far from the decision-boundary (faraway  OODs) can degrade SSL performance substantially if the model contains a batch normalization (BN) layer. The last observation makes sense logically as well since the batch normalization heavily depends on the mean and variance of each batch's data points, which can be significantly different for OOD points that came from very different distributions. 
We find these observations about OOD points consistent across experiments on several synthetic and real-world datasets.

\noindent \textbf{Weighted Robust SSL Framework.} 
Our second contribution is a unified, weighted robust SSL approach to improve many existing SSL algorithms' robustness by learning to assign weights to unlabeled examples based on a bi-level optimization approach.
To address the limitation of low-order approximations in bi-level optimization (DS3L), we designed an implicit-differentiation based algorithm that considered high-order approximations of the objective and is scalable to a higher number of inner optimization steps to learn a massive amount of weight parameters. In addition to address the BN issue due to the faraway OODs, we propose \emph{weighted batch normalization (WBN)} to carry the weights (learned from bi-level optimization) over in the BN step  (Sec.~\ref{sec:WBN}). 

\noindent \textbf{Comprehensive experiments. } We conduct extensive experiments on synthetic and real-world datasets. The results demonstrate that our weighted robust SSL approach significantly outperforms existing robust approaches (L2RW, MWN, Safe-SSL, and UASD) on four representative SSL algorithms. We also perform an ablation study to demonstrate which components of our approach are most important for its success. 


\section{Semi-Supervised Learning (SSL)}
Given a training set with a  labeled set of examples $\mathcal{D} = \{\x_i, y_i\}_{i=1}^n$ and an unlabeled set of examples $\mathcal{U} = \{\x_j\}_{j=1}^m$. For any classifier model $f(\x, \theta)$ used in SSL, where $\x\in \mathbb{R}^C$ is the input data, and $\theta$ refers to the parameters of the classifier model. The loss functions of many existing methods can be formulated as the following general form:
\vspace{-1mm}
\begin{eqnarray}
\sum\nolimits_{(\x_i, y_i)\in \mathcal{D}} l(f(\x_i, \theta), y_i) + \sum\nolimits_{x_j\in \mathcal{U}} r(f(\x_j, \theta)), 
\label{ssl_loss}
\end{eqnarray}
where $l(\cdot)$ is the loss function for labeled data (such as cross-entropy), and $r(\cdot)$ is the loss function (regularization function) on the unlabeled set.  The goal of SSL methods is to design an efficient regularization function to leverage the model performance information on the unlabeled dataset for effective training.
Pseudo-labeling~\cite{lee2013pseudo} uses a standard supervised loss function on an unlabeled dataset using “pseudo-labels” as a target label as a regularizer.
$\Pi$-Model~\cite{laine2016temporal, sajjadi2016regularization} designed a consistency-based regularization function that pushes the distance between
the prediction for an unlabeled sample and its stochastic perturbation (e.g., data augmentation or dropout~\cite{srivastava2014dropout}) to a small value. 
Mean Teacher~\cite{tarvainen2017mean} proposed to obtain a more stable target output $f(x, \theta)$ for unlabeled set by setting the target via an exponential moving average of parameters from previous training steps.
Instead of designing a stochastic $f(x, \theta)$, Virtual Adversarial Training (VAT)~\cite{miyato2018virtual} proposed to approximate a tiny perturbation to unlabeled samples that affect the output of the prediction function most.  MixMatch~\cite{berthelot2019mixmatch}, UDA~\cite{xie2019unsupervised}, and Fix-Match~\cite{sohn2020fixmatch} choose the pseudo-labels based on predictions of augmented samples, such as shifts, cropping, image flipping, weak and strong augmentation, and mix-up~\cite{zhang2017mixup} to design the regularization functions. However, the performance of most existing SSL can degrade substantially when the unlabeled dataset contains OOD examples~\cite{oliver2018realistic}.

\section{Impact of OOD on SSL Performance}\label{sec:Impact_OOD}

In this section, we provide a systematic analysis of the impact of OODs for many popular SSL algorithms, such as Pseudo-Label(PL)~\cite{lee2013pseudo}, $\Pi$-Model~\cite{laine2016temporal}, 
Mean Teacher(MT)~\cite{tarvainen2017mean}, and Virtual Adversarial Training (VAT)~\cite{miyato2018virtual}. We illustrate the discoveries using the following synthetic and real-world datasets. While we mainly focus on VAT as the choice of the SSL algorithm, the observations extend to other SSL algorithms as well. 

\begin{figure*}[!t]
    \centering
    \begin{subfigure}[b]{0.195\textwidth}
        \centering
        \includegraphics[width=\linewidth]{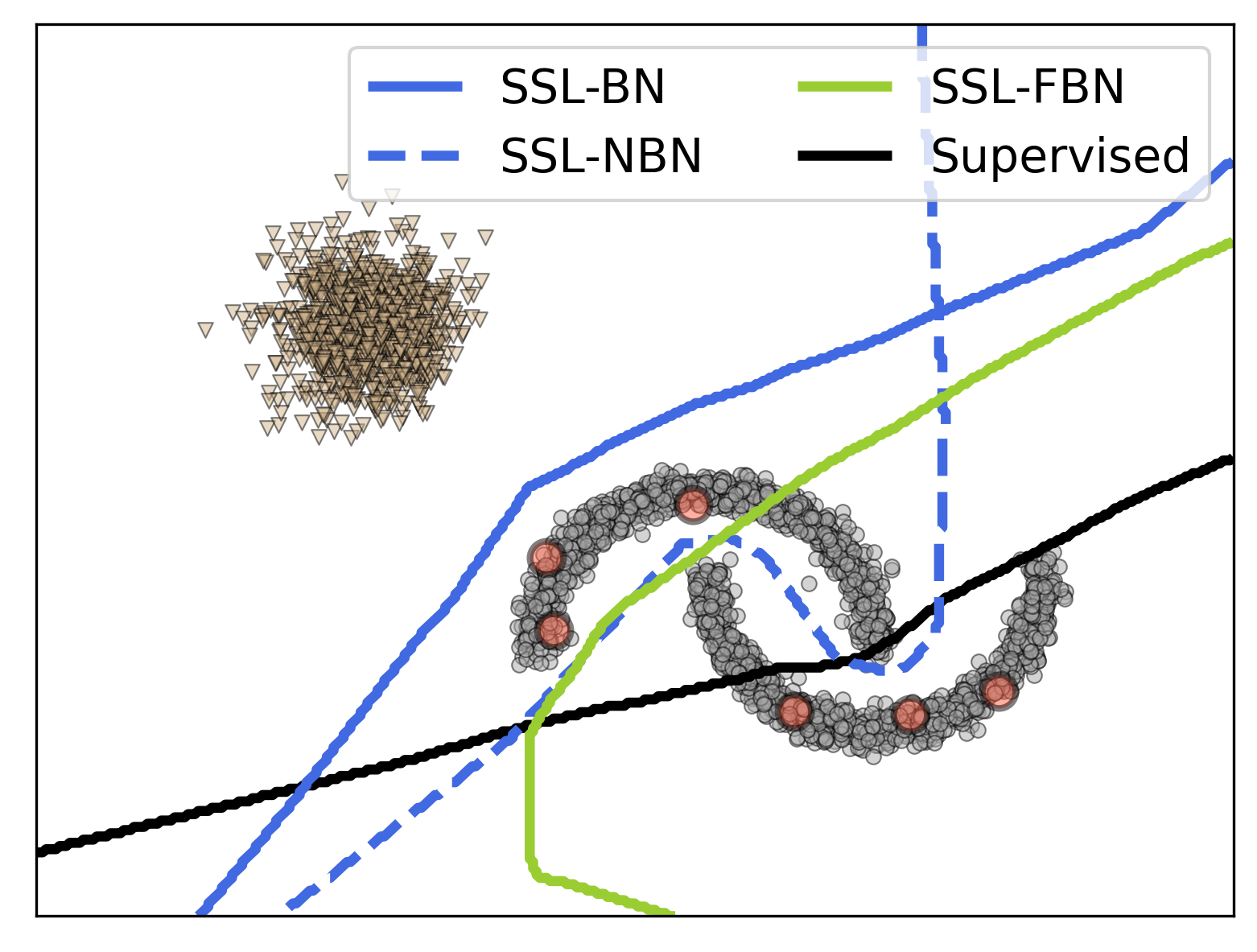}
        \caption{Faraway OODs}
    \end{subfigure}
    \begin{subfigure}[b]{0.195\textwidth}
        \centering
        \includegraphics[width=\linewidth]{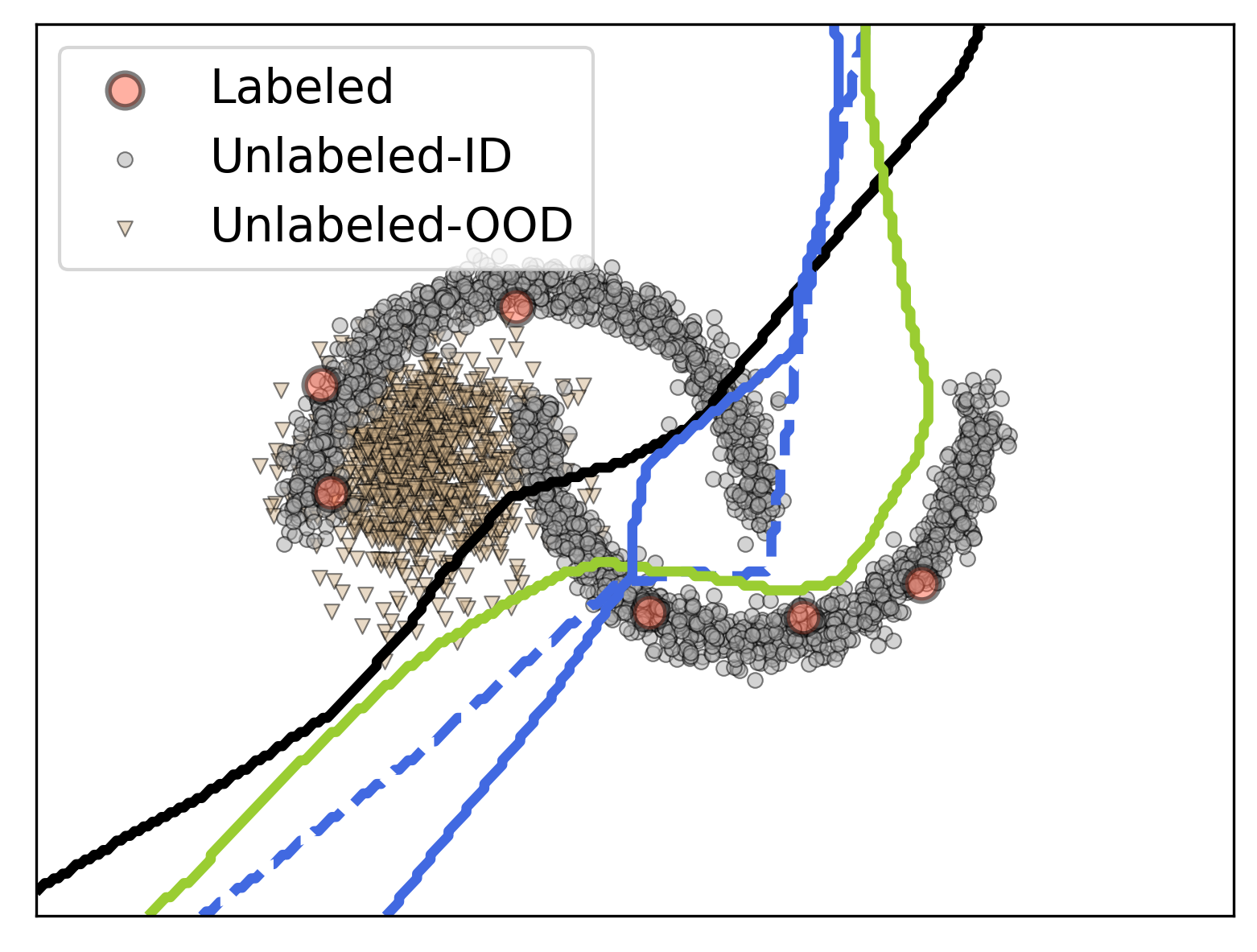}
        \caption{Boundary OODs}
    \end{subfigure}
    \begin{subfigure}[b]{0.195\textwidth}
        \centering
        \includegraphics[width=\linewidth]{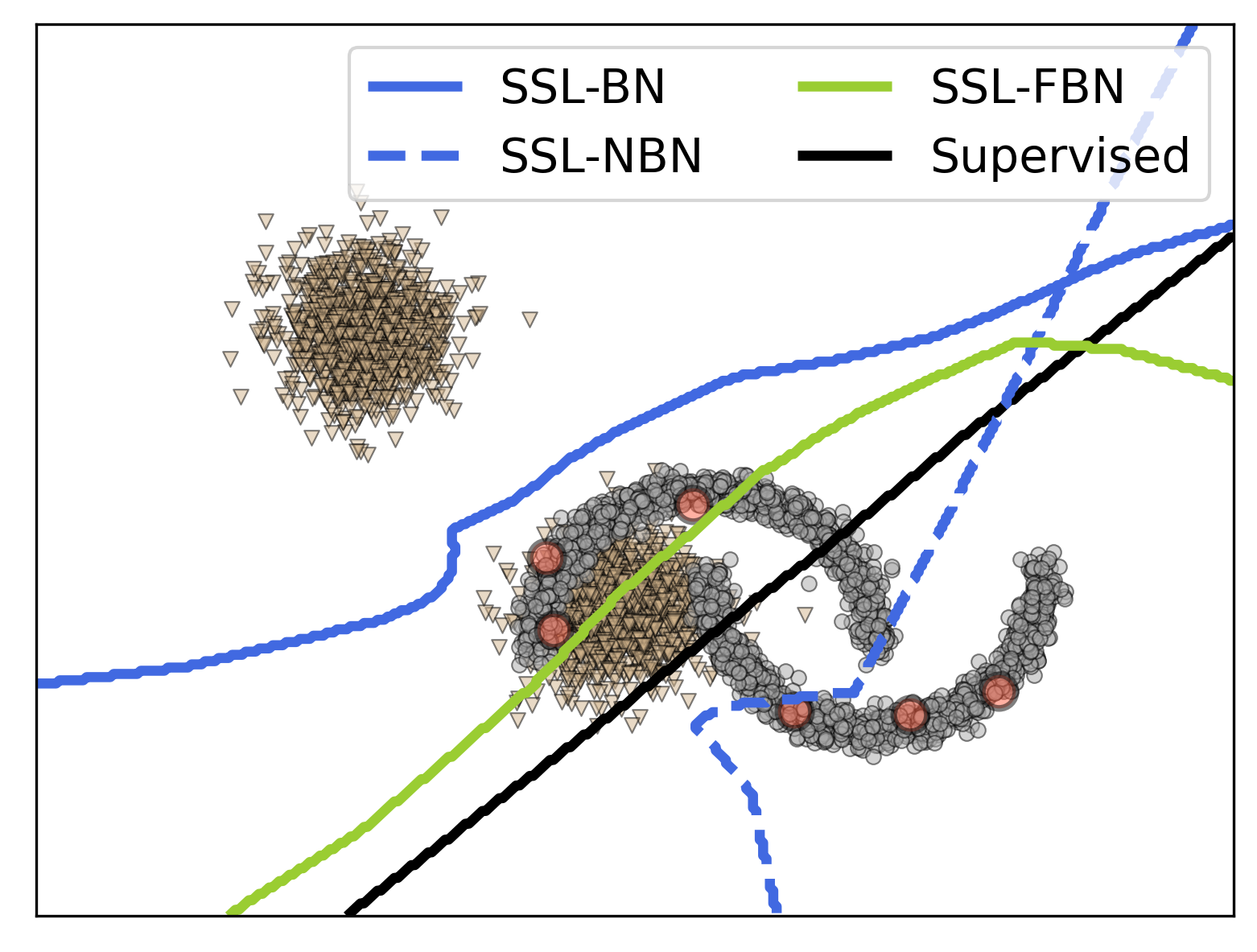}
        \caption{Mixed OODs}
    \end{subfigure}
    \begin{subfigure}[b]{0.195\textwidth}
        \centering
        \includegraphics[width=\linewidth]{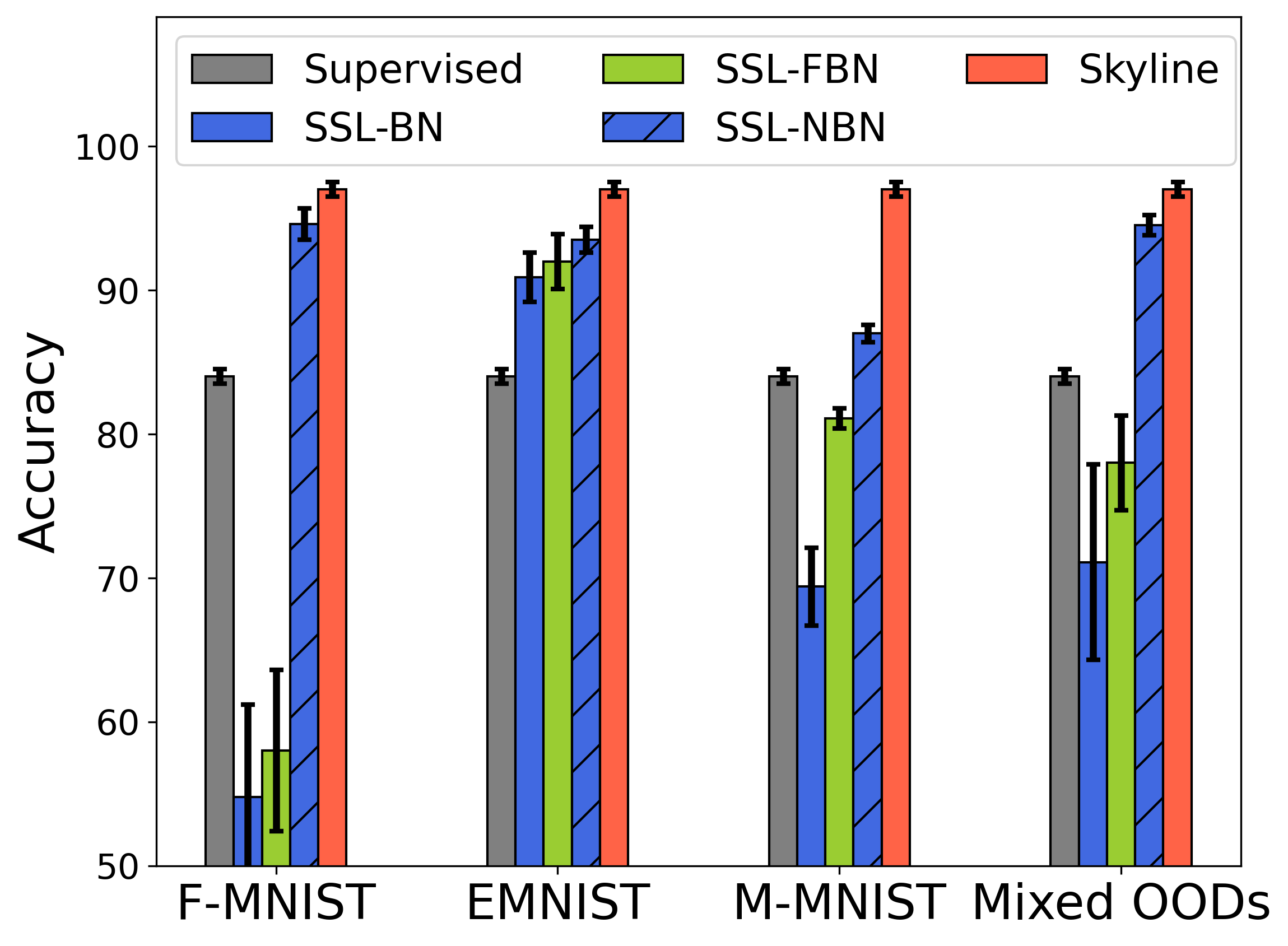}
        \caption{OODs in MNIST}
    \end{subfigure}
     \begin{subfigure}[b]{0.195\textwidth}
        \centering
        \includegraphics[width=\linewidth]{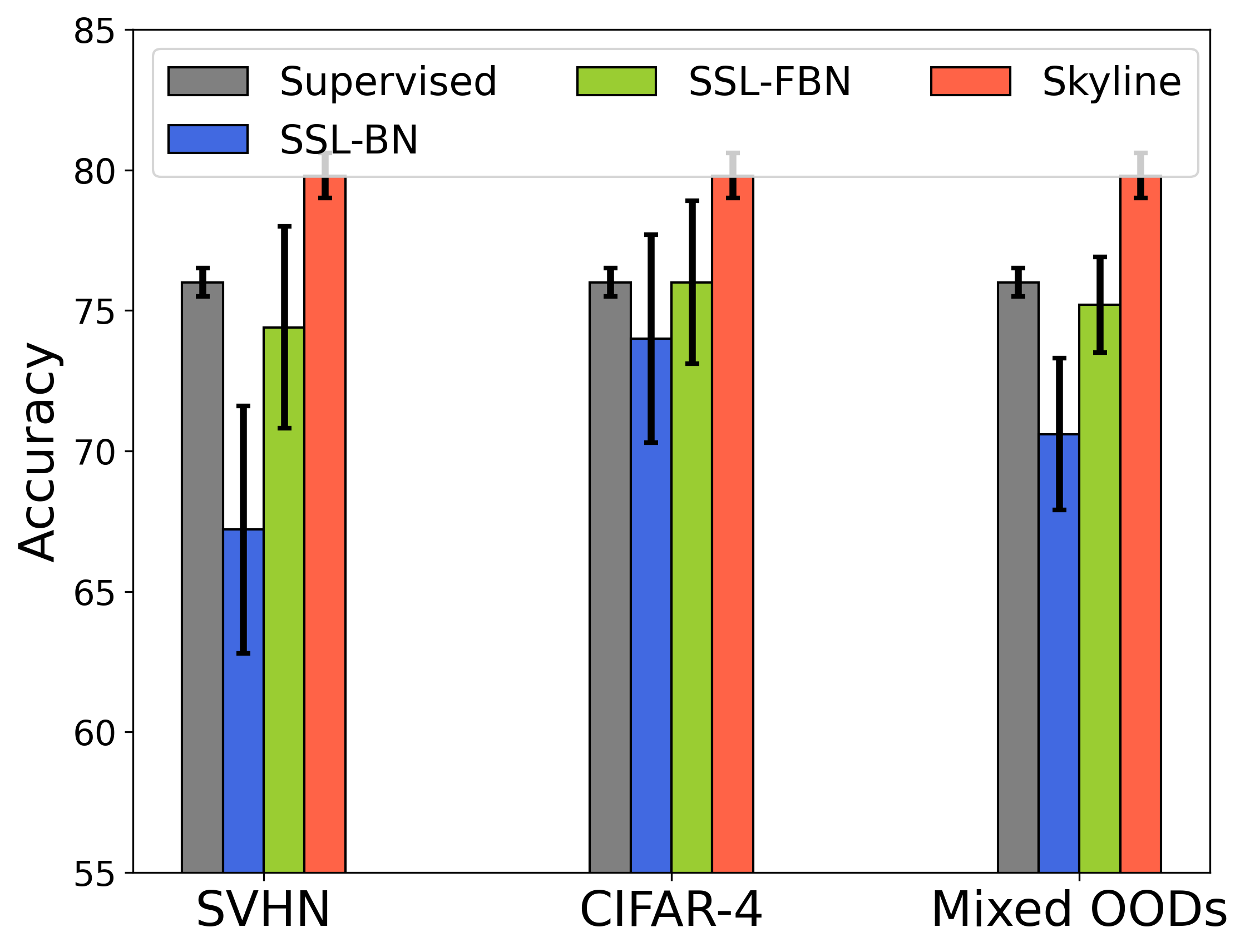}
        \caption{OODs in CIFAR10}
    \end{subfigure}
    \caption{SSL performance with different type OODs in both synthetic and real-world datasets. }
    \label{fig: impact_ood}
    \vspace{-6mm}
\end{figure*}

\noindent {\bf Synthetic dataset.} We considered two moons dataset (red points are labeled data, gray circle points are in-distribution (ID) unlabeled data) with OOD (yellow triangle points)  points in three different scenarios that can exist in real-world, 1) Faraway OOD scenario where the OOD points exist far from decision boundary; 2) Boundary OOD scenario where the OOD points occur close to decision boundary; 3) Mixed OOD scenario where OOD points exist both far and close to the decision boundary, as shown in Fig~\ref{fig: impact_ood}.


\noindent {\bf Real-world dataset.} We consider MNIST as ID data with three types of OODs to account for plausible real-world scenarios. 1)Faraway OOD: We used Fashion MNIST (F-MNIST) dataset, which contains fashion images as Faraway OOD dataset as it inherently has different patterns compared to MNIST dataset; 2) Boundary OOD: We used EMNIST  dataset, which contains handwritten character digits as Boundary OOD dataset as it has similar patterns compared to MNIST dataset; In addition to EMNIST, we also considered Mean MNIST (M-MNIST) as a boundary OOD dataset, which was generated by averaging MNIST images from two different classes (usage of M-MNIST as boundary OOD is also considered in ~\cite{guo2019mixup}); 4) Mixed OOD: For Mixed OOD dataset, we combined both Fashion MNIST and EMNIST together.

For all experiments in this section, we used a multilayer perceptron neural network (MLP) with three layers as a backbone architecture for the synthetic dataset and LeNet as a backbone for the real-world datasets. We consider the following models in the experiments: 1) \emph{SSL-NBN}: MLP or LeNet model without Batch Normalization; 2) \emph{SSL-BN}: MLP or LeNet model with Batch Normalization; 3) \emph{SSL-FBN}: MLP or LeNet model where we freeze the batch normalization layers for the unlabeled instances. Freezing BN (FBN) ~\cite{oliver2018realistic} is a common trick to improve the SSL model robustness where we freeze batch normalization layers by not updating \textit{running\_mean} and \textit{running\_variance} in the training phase.

The following are the main observations. First, from Fig~\ref{fig: impact_ood}, we see that with BN (i.e., SSL-BN), there is a significant impact on model performance and learned decision boundaries in the presence of OOD. This performance degradation is even more pronounced in Faraway OOD since the BN statistics like the running mean/variance can be significantly changed by faraway OOD points. Secondly, when we do not use BN (i.e., SSL-NBN), the impact of the Faraway OOD and mixed OOD data is reduced. However, in the case of boundary OOD (Fig \ref{fig: impact_ood} (b) and EMNIST/M-MNIST case of Fig~\ref{fig: impact_ood} (d)), we still see significant performance degradation compared to the skyline. However, BN is a crucial component in more complicated models (Eg: ResNet family), and we expect OOD instances to play a significant role there. Finally, when freezing the BN layers for the unlabeled data (i.e., SSL-FBN), we see that the Faraway and Mixed OODs' effect is alleviated; but SSL-FBN still performs worse than the SSL-NBN in Faraway and Mixed OODs (and there is big scope of improvement w.r.t the skyline). Finally, both SSL-NBN and SSL-FBN fail to efficiently mitigate the performance degradation caused by boundary OOD data points. We also show that similar observations made on the CIFAR-10 dataset (Fig~\ref{fig: impact_ood} (e)). 

\begin{restatable}{proposition}{primepropositionone1}\label{proposition1}
Give in-distribution mini-batch $\mathcal{I}=\{\x_i\}_{i=1}^m$, OOD mini-batch $\mathcal{O}=\{\hat{\x}_i\}_{i=1}^m$, and the mixed mini-batch $\mathcal{IO}=\mathcal{I}\cup \mathcal{O}$. Denote $\mu_{\mathcal{M}}$ is the mini-batch mean of $\mathcal{M}$ (either $\mathcal{O}, \mathcal{I}$ or $\mathcal{IO}$).
With faraway OOD: $\|\mu_{\mathcal{O}}- \mu_{\mathcal{I}}\|_2 > L $, where $L$ is large ($L\gg0$), we have:
\begin{equation}
\|\mu_{\mathcal{IO}} - \mu_{\mathcal{I}}\|_2 > \frac{L}{2} \quad  \text{and} \quad
    BN_{\mathcal{I}}(\x_i) \neq BN_{\mathcal{IO}}(\x_i) \nonumber
\end{equation}
where $BN_{\mathcal{M}}(\x_i)$ is traditional batch normalizing transform based on mini-batch from $\mathcal{M}$ (either $\mathcal{I}$ or $\mathcal{IO}$).
\end{restatable}

\begin{proof}
The mini-batch mean of $\mathcal{I}$: $ \mu_{\mathcal{I}} = \frac{1}{m}\sum_{i=1}^m \x_i$.
The mixed mini-batch mean of $\mathcal{IO}$: $\mu_{\mathcal{IO}} = \frac{1}{2m}(\sum_{i=1}^m \x_i +\sum_{i=1}^m \hat{\x}_i) 
     = \frac{1}{2}\mu_{\mathcal{I}} + \frac{1}{2}\mu_{\mathcal{O}}.$
Then we have, 
\begin{equation}
    \|\mu_{\mathcal{IO}}- \mu_{\mathcal{I}}\|_2 = \|\frac{1}{2}\mu_{\mathcal{I}} + \frac{1}{2}\mu_{\mathcal{O}} -\mu_{\mathcal{I}}\|_2 >\frac{L}{2} \gg 0 \nonumber
\end{equation}
The mini-batch variance of $\mathcal{I}$: $\sigma^2_{\mathcal{I}} = \frac{1}{m}\sum_{i=1}^m (\x_i - \mu_{\mathcal{I}})^2$.
The traditional batch normalizing transform based on mini-batch $\mathcal{I}$ for $\x_i$: $BN_{\mathcal{I}}(\x_i) = \gamma \frac{\x_i-\mu_\mathcal{I}}{\sqrt{\sigma^2_{\mathcal{I}}+\epsilon}} +\beta
    \label{BN_I}$.
The mini-batch variance of $\mathcal{IO}$,
\begin{eqnarray}
     \sigma^2_{\mathcal{IO}}
     &=& \frac{1}{2m}(\sum_{i=1}^m \x_i^2+ \sum_{i=1}^m \hat{\x}_i^2) -\mu_{\mathcal{IO}}^2 \nonumber \\
     &=& \frac{1}{2}\sigma^2_{\mathcal{I}} + \frac{1}{2}\sigma^2_{\mathcal{O}}+ \frac{1}{4}(\mu_{\mathcal{O}}-\mu_{\mathcal{I}})^2 \nonumber \\
     &\approx& \frac{1}{4}(\mu_{\mathcal{O}}-\mu_{\mathcal{I}})^2
     \label{var_IO}\feng{,}
\end{eqnarray}
and the traditional batch normalizing transform based on mini-batch $\mathcal{IO}$ for $\x_i$,
\begin{eqnarray}
 BN_{\mathcal{IO}}(\x_i) &=& \gamma \frac{\x_i -\mu_{\mathcal{B}}(IO)}{\sqrt{\sigma^2_{\mathcal{B}}(IO)+\epsilon}} +\beta \nonumber \\
 &=& \gamma\Big( \frac{\x_i -\mu_O}{2\sqrt{\sigma^2_{\mathcal{B}}(IO)+\epsilon}} + \frac{\x_i -\mu_I}{2\sqrt{\sigma^2_{\mathcal{B}}(IO)+\epsilon}}\Big) + \beta \nonumber \\
 &\approx& \gamma\Big( \frac{\x_i -\mu_O}{\|\mu_O-\mu_I \|_2} + \frac{\x_i -\mu_I}{\|\mu_O-\mu_I \|_2}\Big) + \beta \nonumber \\
 &\approx& \gamma \frac{\x_i -\mu_O}{\|\mu_O-\mu_I \|_2} + \beta
  \label{BN_IO}.
\end{eqnarray}
The approximations in Eq.~\eqref{var_IO} and Eq.~\eqref{BN_IO} hold when $\sigma^2_{\mathcal{I}}$ and $\sigma^2_{\mathcal{O}}$ have the same magnitude levels as $\mu_I$, i.e., $\|\mu_{\mathcal{O}} - \sigma^2_{\mathcal{I}}\|_2>L$ and $\|\mu_{\mathcal{O}} - \sigma^2_{\mathcal{O}}\|_2 > L$. Comparing Eq.~\eqref{BN_I} with Eq.~\eqref{BN_IO}, we prove that $BN_{\mathcal{I}}(\x_i) \neq BN_{\mathcal{IO}}(\x_i)$. 
\end{proof}

Proposition~\ref{proposition1} shows that when unlabeled set contains OODs, the traditional BN behavior could be problematic, therefore resulting in incorrect statistics estimation, e.g., the output of BN for mixed mini-bath $BN_{\mathcal{IO}}(\x_i) \approx \gamma \frac{\x_i -\mu_{\mathcal{O}}}{\|\mu_{\mathcal{O}}-\mu_{\mathcal{I}} \|_2} + \beta$, which is not our expected result ($BN_{\mathcal{I}}(\x_i) = \gamma \frac{\x_i -\mu_{\mathcal{I}}}{\sqrt{\sigma^2_{\mathcal{I}}}+\epsilon} + \beta$).

The main takeaways of the synthetic and real data experiments are as follows: 1) OOD instances close to decision boundary (Boundary OODs) hurt SSL performance irrespective of the use of batch normalization; 2) OOD instances far from the decision boundary (Faraway OODs) hurt the SSL performance if the model involves BN. Freezing BN can reduce some impact of OOD to some extent but not entirely; 3) OOD instances far from the decision boundary will not hurt SSL performance if there is no BN in the model. To this end, we answered the question "\textit{How out-of-
distribution data hurt semi-supervised learning performance?}".
In the next section, we propose a weighted robust SSL framework to address above mentioned issues caused by OOD data points.

\section{Methodology}
In this section, we first proposed the weighted robust SSL (WR-SSL) framework in Sec.~\ref{sec.4.1}, and then introduce the implicit-differentiation (high-order approximation) based optimization algorithms to train the weighted robust SSL approach in Sec.~\ref{sec.4.2}. More importantly, we proposed weighted batch normalization to improve the robustness of our WR-SSL framework against OODs in Sec.~\ref{sec:WBN}.

\subsection{Weighted Robust SSL Framework}\label{sec.4.1}
\noindent  {\bf Reweighting the unlabeled data.} Consider the semi-supervised classification problem with training data (labeled $\mathcal{D}$ and unlabeled $\mathcal{U}$) and classifier $f(x ;\theta)$. Generally, the optimal classifier parameter $\theta$ can be extracted by minimizing the SSL loss (Eq.~\eqref{ssl_loss}) calculated on the training set. In the presence of unlabeled OOD data, sample reweighting methods enhance the robustness of training by imposing weight $w_j$ on the $j$-th unlabeled sample loss,
\begin{eqnarray}
\mathcal{L}_{T}(\theta, \w) = \sum_{(\x_i, y_i)\in \mathcal{D}} l(f(\x_i; \theta), y_i) + \sum_{x_j\in \mathcal{U}} w_j r(f(\x_j; \theta)), \nonumber
\label{dss_ssl} 
\end{eqnarray}
where we denote $\mathcal{L}_U$ is the robust unlabeled loss, and we treat weight $\w$ as hyperparameter. Our goal is to learn a sample weight vector $\w$ such that $\w=0$ for OODs, $\w=1$ for In-distribution (ID) sample.

\begin{figure*}[!t]
    \centering
    \includegraphics[width=0.72\textwidth]{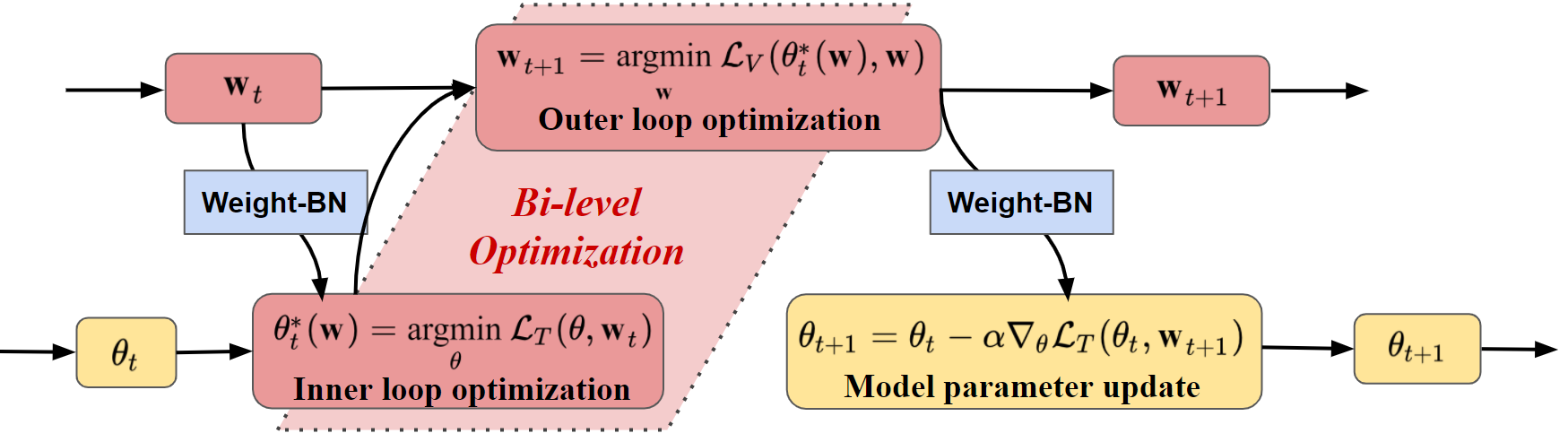}
    \vspace{-2mm}
    \caption{ Main flowchart of the proposed Weighted Robust SSL algorithm.}
    \label{fig:framework}
    \vspace{-5mm}
\end{figure*}

\noindent Denote  $\mathcal{L}_{V}(\theta^*(\w), \w) = \mathcal{L}_{V}(\theta^*(\w))+\lambda \cdot \text{Reg}(\w)$ as the validation loss with a regularization term over the validation dataset, where $\text{Reg}(\w)$ is the regularization term, $\lambda$ is the regularization coefficient, and $\mathcal{L}_{V} (\theta^*(\w)) \triangleq \sum_{(\x_i, y_i)\in \mathcal{V}} l(f(\x_i, \theta^*(\w)), y_i)$. This labeled set could either be a held out validation set, or the original labeled set $\mathcal D$. Intuitively, the problem given in Eq.~\eqref{ssl+dss_w} aims to choose weights of unlabeled samples $\w$ that minimizes the supervised loss evaluated on the validation set when the model parameters $\theta^*(\w)$ are optimized by minimizing the weighted SSL loss $\mathcal{L}_T(\theta, \w)$.

\noindent  {\bf Bi-level optimization objective.} Since manual tuning and grid-search for each $w_i$ is intractable, we pose the weights optimization problem described above as a \emph{bi-level} optimization problem.
\begin{eqnarray}
\vspace{-3mm}
&\mathop{\min}\limits_{\w}& \mathcal{L}_{V}(\theta^*(\w), \w) , \nonumber \\
&\text{s.t. }& \theta^*(\w) = \argmin_{\theta}\mathcal{L}_{T}(\theta, \w). 
\label{ssl+dss_w}
\end{eqnarray}
Calculating the optimal $\theta^*$ and $\w$ requires two nested loops of optimization, which is expensive and intractable to obtain the exact solution, especially when optimization involves deep learning model and large datasets. Since gradient-based methods like Stochastic Gradient Descent (SGD) have shown to be very effective for machine learning and deep learning problems, we adapt a 
high-order approximation strategy, as described in Sec~\ref{sec.4.2}.

\subsection{high-order Optimization Approximation}\label{sec.4.2}
In this section, we developed an efficient high-order optimization algorithms to train our weighted robust SSL framework.

\noindent  {\bf Implicit Differentiation.} Directly calculate the weight gradient $\frac{\partial \mathcal{L}_{V}(\theta^*(\w), \w)}{ \partial \w}$ by chain rule:
\begin{eqnarray}
    \frac{\partial \mathcal{L}_{V} (\theta^*(\w), \w)}{ \partial \w}
    =\underbrace{\frac{\partial \mathcal{L}_{V} }{ \partial \w}}_{(a)} + \underbrace{\frac{\partial \mathcal{L}_{V} }{ \partial \theta^*(\w)}}_{(b)} \times \underbrace{\frac{\partial \theta^*(\w)}{\partial \w}}_{(c)} \label{ID_gradient}
\end{eqnarray}
where (a) is the weight direct gradient (e.g., gradient from regularization term, $\text{Reg}(\w)$), (b) is the parameter direct gradient, which are easy to compute. The difficult part is the term (c) (best-response Jacobian). We approximate (c) by using the Implicit function theorem~\cite{lorraine2020optimizing},
\begin{eqnarray}
    \frac{\partial \theta^* (\w)}{\w} = - \underbrace{\Big[ \frac{\partial \mathcal{L}_{T}}{\partial \theta\partial \theta^T} \Big]^{-1}}_{(d)} \times  \underbrace{\frac{\partial \mathcal{L}_{T}}{\partial \w\partial \theta^T}}_{(e)} \label{IFT}
\end{eqnarray}

However, computing Eq.~\eqref{IFT} is challenging when using deep nets because it requires to invert a high dimensional Hessian (term (d)), which often require $\mathcal{O}(m^3)$ operations. Therefore, we give the Neumann series approximations~\cite{lorraine2020optimizing} of term (d) which we empirically found to be effective for SSL,
\begin{equation}
    \Big[ \frac{\partial \mathcal{L}_{T}}{\partial \theta\partial \theta^T} \Big]^{-1} \approx  \lim_{P \rightarrow{}\infty} \sum_{p=0}^P \Big[I- \frac{\partial \mathcal{L}_{T}}{\partial \theta\partial \theta^T} \Big]^p \label{neumann_serires}
\end{equation}
where $I$ is the identity matrix.

Since the algorithm mentioned in ~\cite{lorraine2020optimizing} utilizes the Neumann series approximation and efficient hessian vector product to compute the Hessian inverse product, it can efficiently compute the hessian-inverse product even when a larger number of weight hyperparameters are present. We should also note that the implicit function theorem's assumption $\frac{\partial \mathcal{L}_{T}}{\partial \theta} = 0$ needs to be satisfied to accurately calculate the Hessian inverse product. However in practice, we only approximate $\theta^{*}$, and simultaneously train both $\w$ and $\theta$ by alternatively optimizing $\theta$ using $\mathcal{L}_T$ and $\w$ using $\mathcal{L}_V$. 

\subsection{Connections to low-order Approximation}

\begin{restatable}{lemma}{primepropositiontwo}\label{implicitlemma}
Suppose that the Hessian inverse of training loss $\mathcal{L}_T$  with the model parameters $\theta$  is equal to the identity matrix 
$\frac{\partial^2 \mathcal{L}_T}{\partial \theta \partial \theta^T} = \mathbf{I}$ (i.e.,$P=0$ for implicit differentiation approach). Suppose the model parameters are optimized using single-step gradient descent (i.e., $J=1$ for the low-order approximation approach), and the model learning rate is equal to one. Then, the weight update step in both implicit differentiation and low-order approximation approach is equal.
\end{restatable}

\begin{proof}
\begin{case} \textbf{Our Approach:}
In Eq.~\ref{neumann_serires}, we have $\Big[ \frac{\partial \mathcal{L}_{T}}{\partial \theta\partial \theta^T} \Big]^{-1} =  \mathbf{I}$ and substituting it in Eq.~\eqref{IFT}, we have:
\begin{eqnarray}
    \frac{\partial \theta^* (\w)}{\w} = - \frac{\partial \mathcal{L}_{T}}{\partial \w\partial \theta^T}
\end{eqnarray}

Substituting the above equation in Eq.~\eqref{ID_gradient}, we have:
\begin{eqnarray}
    \frac{\partial \mathcal{L}_{V} (\theta^*(\w), \w)}{ \partial \w}
    =\frac{\partial \mathcal{L}_{V} }{ \partial \w} - \frac{\partial \mathcal{L}_{V}(\theta^*(\w)) }{ \partial \theta^*(\w)} \times \frac{\partial^2 \mathcal{L}_{T}}{\partial \w\partial \theta^T}
\end{eqnarray}

Since, we are using unweighted validation loss $\mathcal{L}_V$, there is no dependence of validation loss on weights directly i.e., $\frac{\partial \mathcal{L}_{V}}{\partial \w} = 0$. Hence, the weight gradient is as follows:
\begin{eqnarray}
    \frac{\partial \mathcal{L}_{V} (\theta^*(\w), \w)}{ \partial \w} = - \frac{\partial \mathcal{L}_{V}(\theta^*(\w)) }{ \partial^2 \theta^*(\w)} \times \frac{\partial \mathcal{L}_{T}}{\partial \w\partial \theta^T}
\end{eqnarray}

Since we are using one-step gradient approximation, we have $\theta^*(\w) = \theta - \alpha \frac{\partial \mathcal{L}_T(\w, \theta)}{\partial \theta}$ where $\alpha$ is the model parameters learning rate.

The weight update step is as follows:
\begin{eqnarray}
    \w^* = \w + \beta \frac{\partial \mathcal{L}_{V}(\theta^*(\w)) }{ \partial \theta^*(\w)} \times \frac{\partial^2 \mathcal{L}_{T}}{\partial \w\partial \theta^T}
\end{eqnarray}
where $\beta$ is the weight learning rate.
\end{case}

\begin{case} \textbf{low-order approximation Approach} 

In low-order approximation approach, we have $\theta^*(\w) = \theta - \alpha \frac{\partial \mathcal{L}_T(\w, \theta)}{\partial \theta}$ where $\alpha$ is the model parameters learning rate.

Using the value of $\theta^{*}$, the gradient of validation loss with weight hyperparameters is as follows:

\begin{align}
    \frac{\partial \mathcal{L}_{V} (\theta^*(\w))}{ \partial \w} &= \frac{\partial \mathcal{L}_{V}(\theta - \alpha \frac{\partial \mathcal{L}_T(\w, \theta)}{\partial \theta})}{\partial \w} \nonumber \\
    &= - \frac{\partial \mathcal{L}_{V}(\theta^*(\w))}{\partial \theta^*(\w)} \times \alpha \frac{\partial^2 \mathcal{L}_T(\w, \theta)}{\partial \theta \partial \w^T}
\end{align}

Assuming $\alpha = 1$, we have:
\begin{align}
    \frac{\partial \mathcal{L}_{V} (\theta^*(\w))}{ \partial \w} = - \frac{\partial \mathcal{L}_{V}(\theta^*(\w))}{\partial \theta^*(\w)} \times \frac{\partial^2 \mathcal{L}_T(\w, \theta)}{\partial \theta \partial \w^T}
\end{align}

Hence, the weight update step is as follows:
\begin{eqnarray}
    \w^* = \w + \beta \frac{\partial \mathcal{L}_{V}(\theta^*(\w)) }{ \partial \theta^*(\w)} \times \frac{\partial^2 \mathcal{L}_{T}}{\partial \w\partial \theta^T}
\end{eqnarray}
where $\beta$ is the weight learning rate
\end{case}
As shown in the cases above, we have a similar weight update. Hence, it inherently means that using the low-order approximation of $J=1$ is equivalent to using an identity matrix as the Hessian inverse of training loss with $\theta$.
\end{proof}

Lemma~\ref{implicitlemma} shows that the weight gradient from implicit differentiation is same as the weight gradient from the low-order approximation method (e.g., DS3L) when $P=0, J=1$.

\subsection{Weighted Batch Normalization} \label{sec:WBN}
In practice, most deep SSL model would use deep CNN. While BN usually serves as an essential component for many deep CNN models~\cite{he2016deep}. Specifically, BN normalizes input features by the mean and variance computed within each mini-batch. At the same time, OODs would indeed affect the SSL performance due to BN (discussed this issues in Sec.~\ref{sec:Impact_OOD}). 
To address this issue, we proposed a \textit{Weighted Batch Normalization}  (WBN) that performs the normalization for each training mini-batch with sample weights $\w$. We present the WBN in Algorithm~\ref{algorithm_WBN}, where $\epsilon$ is a constant added to the mini-batch variance for numerical stability.

\begin{restatable}{proposition}{primepropositionone}\label{proposition3}
With faraway OOD: $\|\mu_{\mathcal{O}}- \mu_{\mathcal{I}}\|_2 > L $, where $L$ is large ($L\gg0$), given perfect weights $\w=\w_\mathcal{I} \cup \w_\mathcal{O}$, where $\w_\mathcal{I}={\bm 1}$ for mini-batch $\mathcal{I}$ and $\w_{\mathcal{O}}={\bm 0}$ for mini-batch $\mathcal{O}$, we have 
\begin{equation}
  \mu_{\mathcal{I}} = \mu^{\w}_{\mathcal{IO}} \quad \text{and} \quad BN_{\mathcal{I}}(\x_i) = WBN_{\mathcal{IO}}(\x_i, \w) 
\end{equation}
where $\mu^{\w}_{\mathcal{IO}}$ is the weighted mini-batch mean of $\mathcal{IO}$, and $WBN_{\mathcal{IO}}(\x_i, \w)$ is weighted batch normalizing transform based on the set $\mathcal{IO}$ with weights $\w$.  
\end{restatable}

\begin{proof}
The symbols $\mathcal{I}, \mathcal{O}, \mathcal{IO}, \mu_{\mathcal{I}}, BN_{\mathcal{M}}(\x_i)$ are introduced in Proposition~\ref{proposition1}. The weighted mini-batch mean of
$\mathcal{IO}$: 
\begin{eqnarray}
     \mu^\w_{\mathcal{IO}} &=& \frac{\sum_{i=1}^m w^i_{\mathcal{I}} \x_i +\sum_{i=1}^m w^i_{\mathcal{O}} \hat{\x}_i}{\sum_{i=1}^m w^i_{\mathcal{I}}+ \sum_{i=1}^m w^i_{\mathcal{O}}}  \nonumber \\
     &=& \frac{\sum_{i=1}^m 1\cdot \x_i +\sum_{i=1}^m 0\cdot \hat{\x}_i}{\sum_{i=1}^m 1+ \sum_{i=1}^m 0} = \mu_{I},
\end{eqnarray}
witch proves that $\mu_{\mathcal{I}} = \mu^{\w}_{\mathcal{IO}}$. The weighted mini-batch variance of $\mathcal{IO}$,
\begin{eqnarray}
     {\sigma^{\w}_{\mathcal{IO}}}^2 &=& \frac{\sum_{i=1}^m w^i_{\mathcal{I}} (\x_i-\mu^{\w}_{\mathcal{IO}})^2}{\sum_{i=1}^m w^i_{\mathcal{I}}+ \sum_{i=1}^m w^i_{\mathcal{O}}}  
     + \frac{\sum_{i=1}^m w^i_{\mathcal{O}} (\hat{\x}_i--\mu^{\w}_{\mathcal{IO}})^2}{\sum_{i=1}^m w^i_{\mathcal{I}}+ \sum_{i=1}^m w^i_{\mathcal{O}}}  \nonumber \\
     &=& \frac{\sum_{i=1}^m 1\cdot (\x_i-\mu_{\mathcal{IO}})^2 }{m}  = \sigma^2_{\mathcal{I}}. \nonumber
\end{eqnarray}
The weighted batch normalizing transform based on mini-batch $\mathcal{IO}$ for $\x_i$: $WBN_{\mathcal{IO}}(\x_i, \w) = \gamma \frac{\x_i-\mu^{\w}_\mathcal{IO}}{\sqrt{{\sigma^{\w}_{\mathcal{IO}}}^2+\epsilon}} +\beta =  \gamma \frac{\x_i-\mu_\mathcal{I}}{\sqrt{\sigma^2_{\mathcal{I}}+\epsilon}} +\beta$
which proves that $BN_{\mathcal{I}}(\x_i) = WBN_{\mathcal{IO}}(\x_i, \w)$.
\end{proof}

Proposition~\ref{proposition3} shows that our proposed weighted batch normalization (WBN) can reduce the OOD effect and get the expected result. Therefore, our weighted robust SSL framework uses WBN instead of BN if model includes BN layer.
Ablation studies in Sec.~\ref{experiment} demonstrates that such our approach with WBN can improve performance further. Finally, our weighted robust SSL framework is detailed in Algorithm~\ref{algorithm_robustssl}.

\vspace{-2mm}
\begin{algorithm}
\small{
\DontPrintSemicolon
\KwIn{A mini-batch $\mathcal{M}=\{\x_i\}_{i=1}^m$ and sample weight $\textcolor{blue}{\w}=\{w_i\}_{i=1}^m$; Parameters to be learned: $\gamma, \beta$}
\KwOut{$\{t_i= \text{\textit{WBN}}_{\mathcal{M}}(\x_i,\textcolor{blue}{\w}) \}$} 
\SetKwBlock{Begin}{function}{end function}
{
  Weighted mini-batch mean:
  $\mu^{\textcolor{blue}{\w}}_{\mathcal{M}}\leftarrow \frac{1}{\sum_{i=1}^m \textcolor{blue}{w_i}} \sum_{i=1}^m \textcolor{blue}{w_i} \x_i  $ \;
  Weighted mini-batch variance:
  $ {\sigma^{\textcolor{blue}{\w}}_{\mathcal{M}}}^2  \leftarrow \frac{1}{\sum_{i=1}^m \textcolor{blue}{w_i}} \sum_{i=1}^m \textcolor{blue}{w_i}(\x_i -\mu^{\textcolor{blue}{\w}}_{\mathcal{M}})^2$ \;
  Normalize:
  $\hat{x}_i \leftarrow \frac{\x_i -\mu_{\mathcal{M}}^{\textcolor{blue}{\w} } }{\sqrt{{\sigma^{\textcolor{blue}{\w}}_{\mathcal{M}}}^2 +\epsilon}}\nonumber$ \;
  Scale and shift:
  $t_i \leftarrow  \gamma \hat{x}_i + \beta \equiv \text{\textit{WBN}}_{\mathcal{M}}(\x_i, \textcolor{blue}{\w})$ \;
}
\caption{\textcolor{blue}{Weighted} Batch Normalization}\label{algorithm_WBN}
}
\end{algorithm}

\vspace{-2mm}
\begin{algorithm}
\small{
\DontPrintSemicolon
\KwIn{Labeled: $\mathcal{D}$, Unlabeled: $\mathcal{U}$, Reg param: $\lambda$}
\KwOut{Model params: $\theta$, Instance/Cluster Weights: $\w$} 
\SetKwBlock{Begin}{function}{end function}
{
  Set $t=0$; learning rate $\alpha$, $\beta$; \;
  Initialize model parameters $\theta$ and weight $\w$; \;
  Apply K-means to $\mathcal{U}$ (if CRW); \;
    Apply WBN instead of BN (if model includes BN); \;
 \Repeat{convergence}
 {
 \textcolor{gray}{$/*$ Inner loop optimization, initial $\theta^0_t = \theta_t$ $*/$ } \;
 \For{$j=1, \ldots, J$}{$ \theta^j_{t} (\w) = \theta^{j-1}_{t} - \alpha \nabla_{\theta} \mathcal{L}_{T}(\theta^{j-1}_{t}, \w_{t})$} 
 \textcolor{gray}{$/*$ Outer loop optimization, set $\theta^*_t = \theta^J_t$ $*/$ }\;
 Approximate inverse Hessian via Eq.~\eqref{neumann_serires};\;
 Calculate best-response Jacobian by Eq.~\eqref{IFT};\;
 Calculate weight gradient $\nabla_{\w}\mathcal{L}_{V}$ via Eq.~\eqref{ID_gradient};\;
 update weight via $\w_{t+1} = \w_{t} - \beta \cdot \nabla_{\w}\mathcal{L}_{V}$;\;
\textcolor{gray}{$/*$ Update net parameters $*/$} \;
$\theta_{t+1} = \theta_{t} - \alpha \nabla_{\theta} \mathcal{L}_{T}(\theta_{t}, \w_{t+1})$\; 
$t=t+1$\;
 }
\Return{$\theta_{t+1}, \w_{t+1}$}
}
\caption{Weighted Robust SSL}\label{algorithm_robustssl}
}
\end{algorithm}
\vspace{-3mm}

\subsection{Additional Implementation Details}
In this subsection, we discuss additional implementational and practical tricks to make our weighted robust SSL scalable and efficient.

\noindent {\bf Last-layer gradients.} Computing the gradients over deep models is time-consuming due to an enormous number of parameters in the model. To address this issue, we adopt a last-layer gradient approximation similar to ~\cite{ash2019deep,killamsetty2021retrieve,killamsetty2021grad} by
only considering the last classification layer gradients of the classifier model in inner loop optimization (step 8 in algorithm~\ref{algorithm_robustssl}). By simply using the last-layer gradients, we achieve significant speedups in weighted robust SSL.

\noindent {\bf Infrequent update $\w$.} We update the weight parameters every $L$ iterations ($L \ge 2$). In our experiments, we see that we can set $L = 5$ without significant loss in accuracy. For MNIST experiments, we can be even more aggressive and set $L = 20$.

\vspace{-1mm}
\noindent  {\bf Weight Sharing and Regularization.} Considering the entire weight vector $\w$ (overall unlabeled points) is not practical for large datasets and easily overfits (see ablation study experiments), we propose two ways to fix this. The first is weight sharing via clustering, which we call Cluster Re-weight (CRW) method. Specifically, we use an unsupervised cluster algorithm (e.g., K-means algorithm) to embed unlabeled samples into $K$ clusters and assign a weight to each cluster such that we can reduce the dimensionality of $\w$ from $|M|$ to $|K|$, where $|K| \ll |M|$. In practice, for high dimensional data, we may use a pre-trained model to calculate embedding for each point before applying the cluster method. In cases where we do not have an effective pre-trained model for embedding, we consider another variant that applies weights to every unlabeled point but considers an L1 regularization in Eq~\eqref{ssl+dss_w} for sparsity in $\w$. We show that both these tricks effectively improve the performance of reweighting and prevent overfitting on the validation set.

\begin{figure*}[!t]
    \centering
    \begin{subfigure}[b]{0.245\textwidth}
        \centering
        \includegraphics[width=\linewidth]{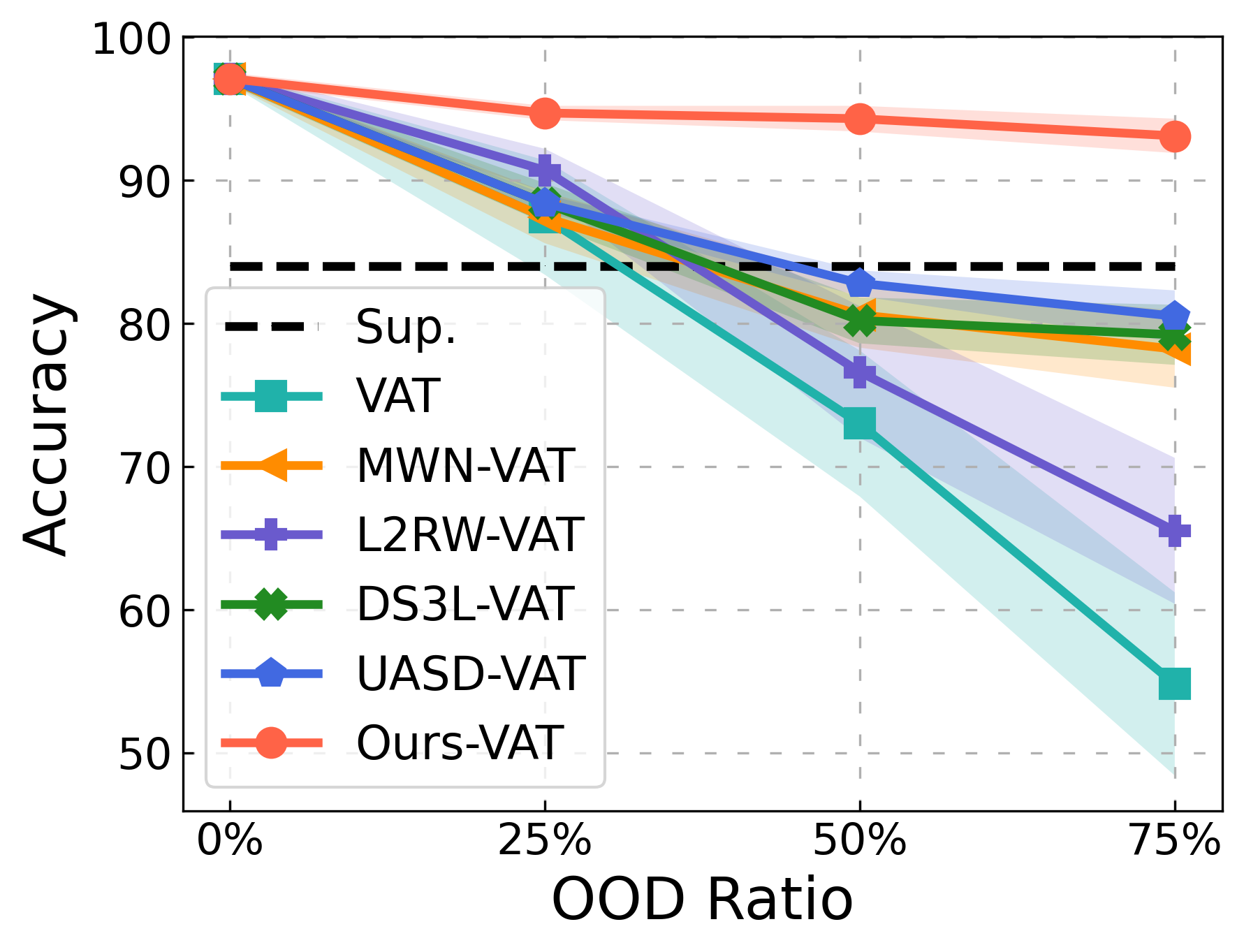}
        \caption{Fashion MNIST.}
        \vspace{-1mm}
    \end{subfigure}
     \hfill
    \begin{subfigure}[b]{0.245\textwidth}
        \centering
        \includegraphics[width=\linewidth]{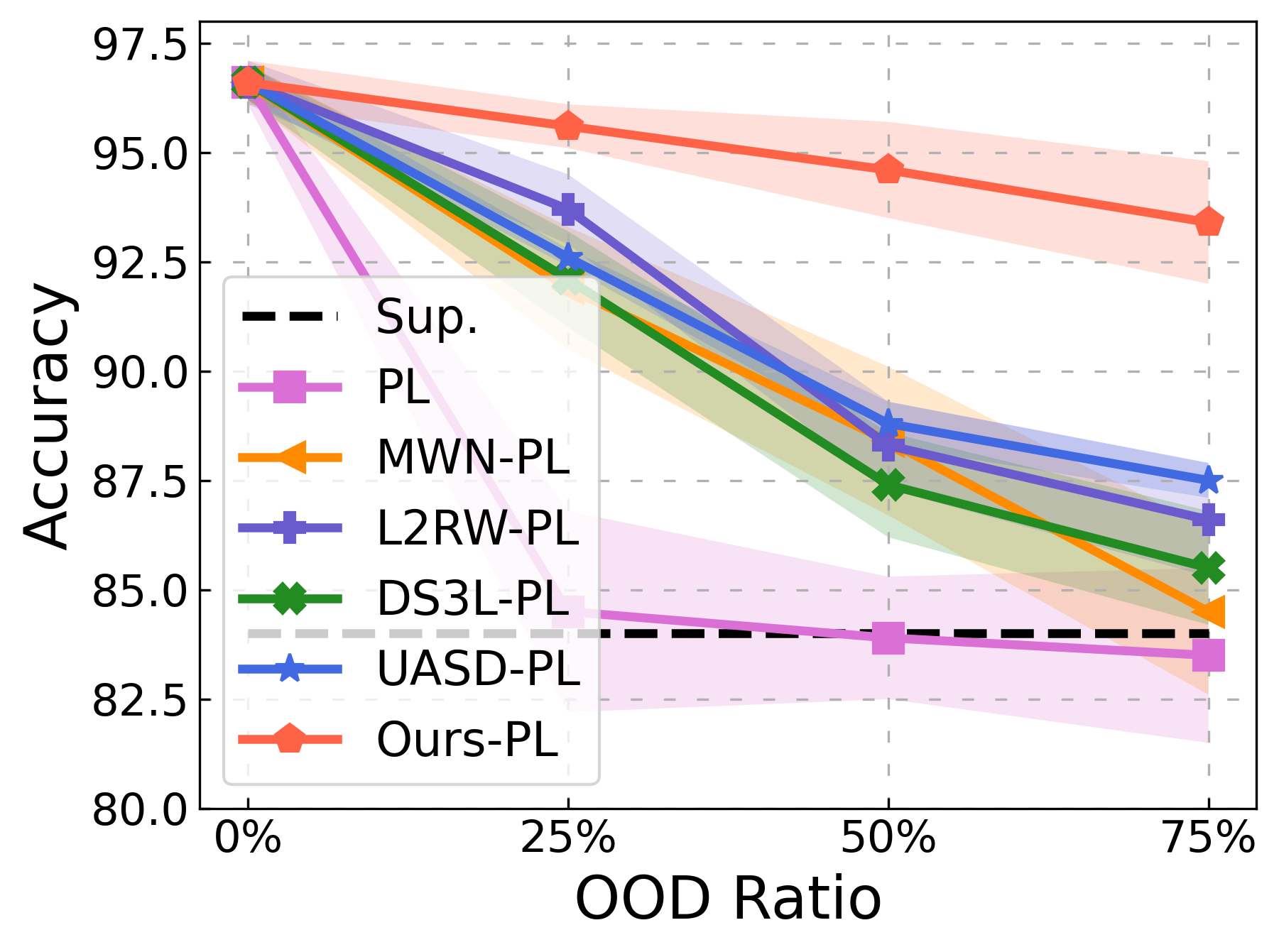}
        \caption{Mean MNIST.}
        \vspace{-1mm}
    \end{subfigure}
    \hfill
    \begin{subfigure}[b]{0.245\textwidth}
        \centering
        \includegraphics[width=\linewidth]{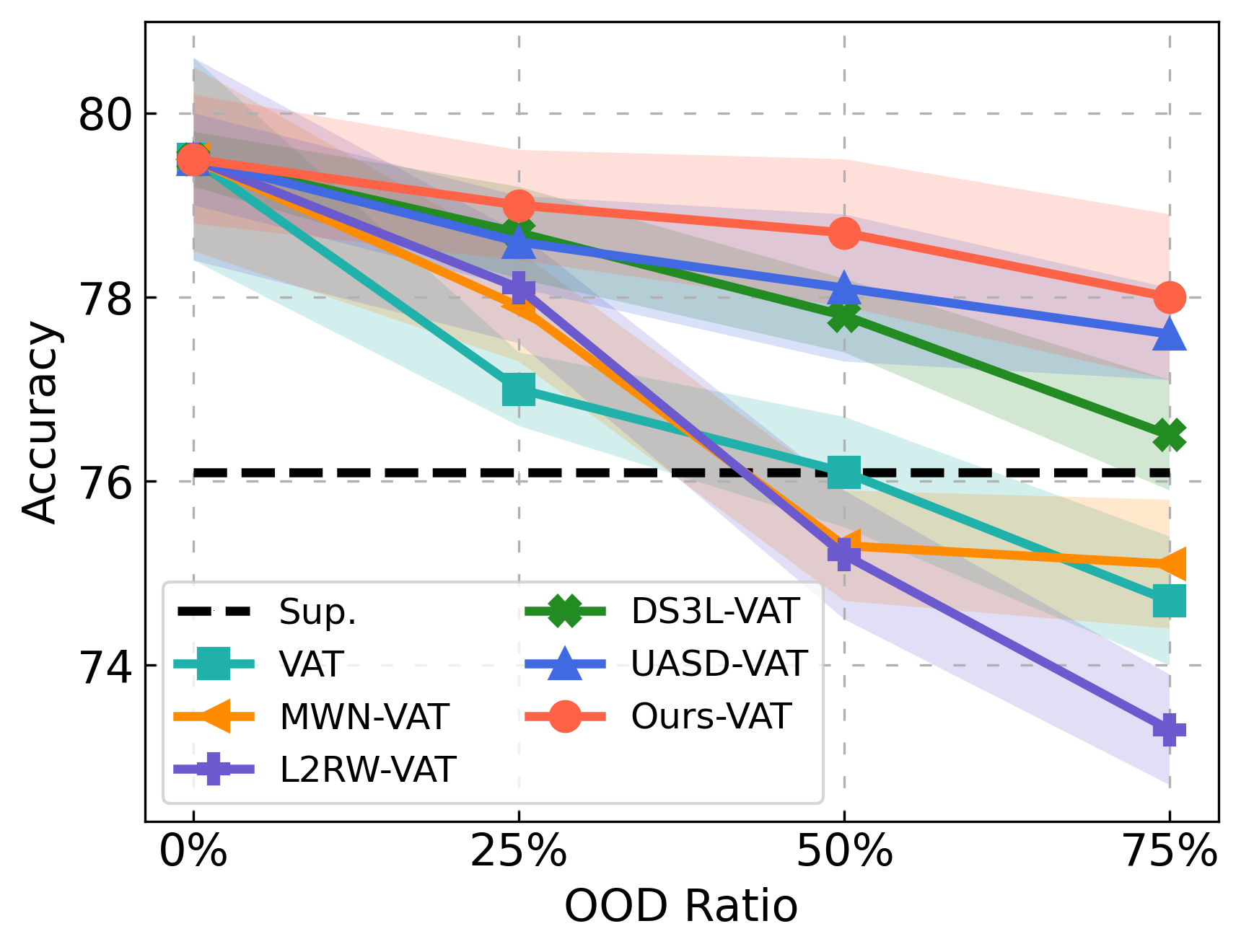}
        \caption{CIFAR10.}
        \vspace{-1mm}
    \end{subfigure}
    \hfill
    \begin{subfigure}[b]{0.245\textwidth}
        \centering
        \includegraphics[width=\linewidth]{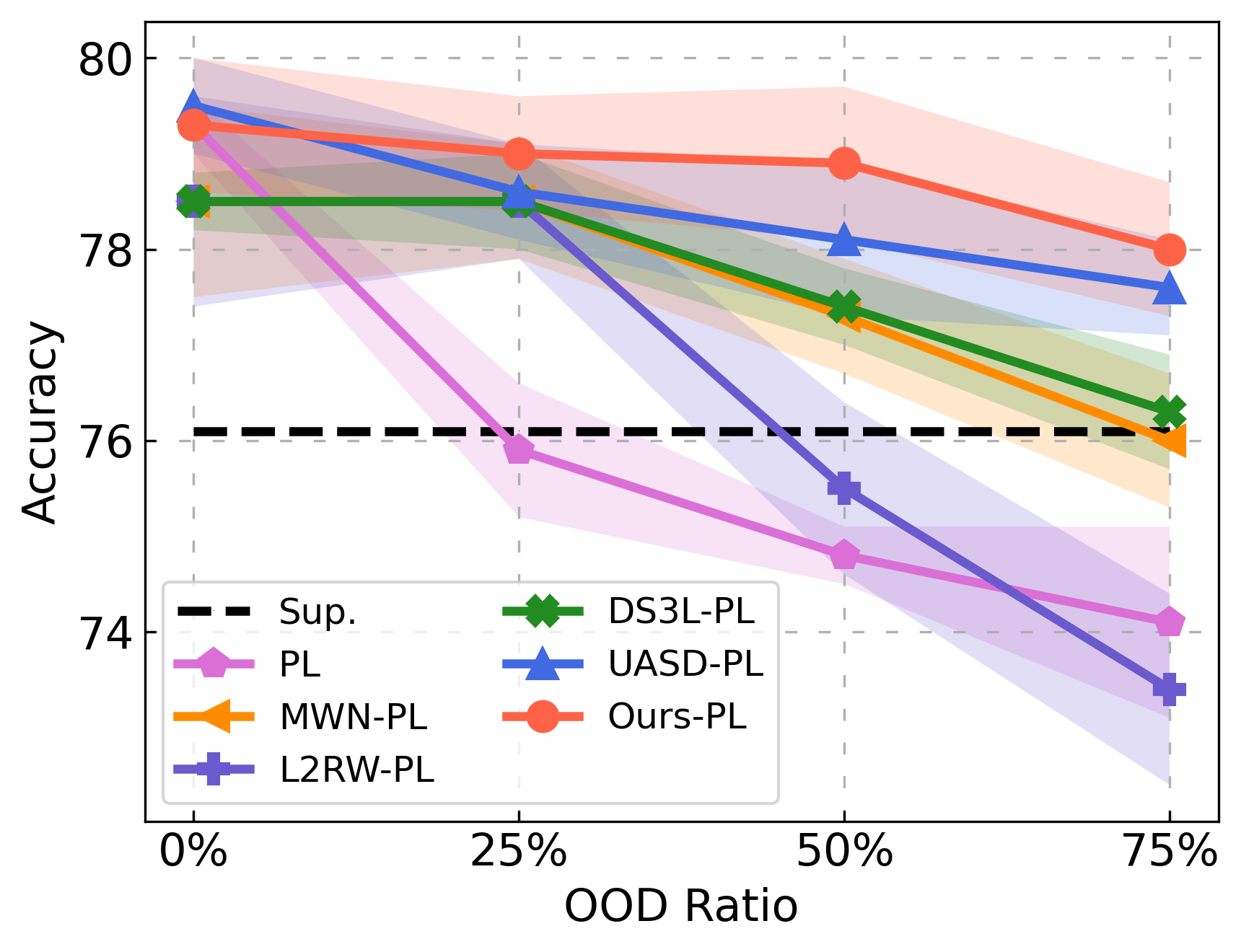}
        \caption{CIFAR10.}
        \vspace{-1mm}
    \end{subfigure}
    \vspace{-4mm}
    \small{
    \caption{(a-d) show test accuracy with varying OOD ratios of 0\% to 75\%. (a) shows results with MNIST as ID and F-MNIST as OOD, (b) shows results with MNIST as ID and M-MNIST as OOD, (c) (d) shows results with CIFAR10 dataset (6 classes ID, 4 classes OOD). (a) (c) (d) uses BN layers for all baselines except our methods that uses WBN layers. (b) shows all methods' results using a model w/o BN layers. (a) (c) use VAT, whereas (b) (d) use PL. Our methods consistently outperform all baselines across different datasets, base SSL algorithms, and models with or without BN layers.}
    \label{fig:4}
    }
\vspace{-5mm}
\end{figure*}

\section{Experimental}\label{experiment}
To corroborate our algorithm, we conduct extensive experiments
comparing our approach (WR-SSL) with some popular baseline methods. We aim to
answer the following questions:

\noindent {\bf Question 1:} Can our approach (WR-SSL) achieve better performance on both different types of OODs with varying OOD ratios compared with baseline methods?

\noindent {\bf Question 2:} How do our approach compare in terms of running time comparing with baseline methods? 

\noindent {\bf Question 3:} What is the effect of each of the components of our approach (e.g., WBN, clustering/regularization, inverse Hessian approximation,  inner loop gradient steps)?  

\subsection{Datasets}
We consider four image classification benchmark datasets. (1)~\textbf{MNIST}: a handwritten digit classification dataset, with
50,000/ 10,000/ 10,000 training/validation/test samples, with training data, split into two groups -- labeled and unlabeled in-distribution (ID) images (labeled data has ten images per class), and with two types of OODs: a) Fashion MNIST, b) Mean MNIST;
(2)~\textbf{CIFAR10}: a natural image dataset with 45,000/ 5,000/ 10,000 training/validation/test samples from 10 object classes, and following~\cite{oliver2018realistic}, we adapt CIFAR10 to a 6-class classification task, using 400 labels per class (from the 6 classes) and rest of the classes OOD (ID classes are "bird", "cat", "deer", "dog", "frog", "horse", and OOD data are from classes: "airline", "automobile", "ship", "truck"); 
(3)~\textbf{CIFAR100}: another natural image dataset with 45,000/5000/10,000 training/validation/test images, similar to CIFAR10, we adapt CIFAR100 to a 50-class classification task, with 40 labels per class -- the ID classes are the first 50 classes, and OOD data corresponds to the last 50 classes;
(4)~\textbf{SVHN-extra} : This is SVHN dataset with 531,131 additional digit images, and we adapt SVHN-extra to a 5-class classification task, using 400 labels per class. The ID classes are the first five classes, and OOD data corresponds to the last five classes.

\vspace{-1mm}
\subsection{Comparing Methods}
\vspace{-1mm}
To evaluate the effectiveness of our proposed weighted robust SSL approaches, 
we compare with five state-of-the-art robust SSL approaches, including UASD~\cite{chen2020semi}, DS3L (DS3L)~\cite{guo2020safe}, L2RW~\cite{ren2018learning}, and MWN~\cite{shu2019meta}. The last two approaches L2RW and MWN, were originally designed for robust supervised learning (SL), and we adapted them to robust SSL by replacing the supervised learning loss function with an SSL loss function. 
We compare these robust approaches on four representative SSL methods, including Pseudo-Label (PL)~\cite{lee2013pseudo}, $\Pi$-Model (PI)~\cite{laine2016temporal,sajjadi2016regularization}, Mean Teacher (MT)~\cite{tarvainen2017mean}, and Virtual Adversarial Training (VAT)~\cite{miyato2018virtual}. One additional baseline is the supervised learning method, named "Sup," which ignored all the unlabeled examples during training. Even Fixmatch~\cite{sohn2020fixmatch} is the state-of-the-art SSL algorithm. We did not consider Fixmatch as the representative SSL method because Fixmatch is non-efficiency (100 hours for training in 1 2080Ti GPU)
All the compared methods were built upon the open-source Pytorch implementation 
by~\cite{oliver2018realistic}. 

\vspace{-1mm}
\subsection{Setup}
\vspace{-1mm}
In our experiments, we implement our approaches (Ours-SSL) for four representative SSL methods, including Pseudo-Label (PL), $\Pi$-Model (PI), Mean Teacher (MT), and Virtual Adversarial Training (VAT). The term "SSL" in Ours-SSL represent the SSL method, e.g., (Ours-VAT denotes our weighted robust SSL algorithm implemented based on VAT.
We used the standard LeNet model as the backbone for the MNIST experiment and used \textit{WRN-28-2} as the backbone for CIFAR10, CIFAR100, and SVHN experiments. For a comprehensive and fair comparison of the CIFAR10 experiment, we followed the same experiment setting of~\cite{oliver2018realistic}. All the compared methods were built upon the Pytorch implementation by ~\cite{oliver2018realistic}. 

\noindent {\bf Hyperparameter setting.} For our WR-SSL approach, we update the weights only using last layer for the inner optimization,  we set $J=3$ (for inner loop gradient steps), $P = 5$ (for inverse Hessian approximation) , $K=20$ (for CRW), $\lambda = 10^{-7}$ (for L1), and $L = 5$ (for infrequent update) for all experiments. We trained all the networks for 2,000 updates with a batch size of 100 for MNIST experiments, and 500,000 updates with a batch size of 100 for CIFAR10, CIFAR100, and SVHN experiments.
We did not use any form of early stopping but instead continuously monitored the validation set performance and report test error at the point of the lowest validation error.
We show the specific hyperparameters used with four representative SSL methods on MNIST experiments in Table~\ref{hyper_mnist}. 
For CIFAR10, we used the same hyperparameters as~\cite{oliver2018realistic}. For CIFAR100 and SVHN datasets, we used the same hyperparameters as CIFAR10\footnote{The source code is accessible at \href{https://github.com/zxj32/WR-SSL}{\color{black}{https://github.com/zxj32/WR-SSL}}}.

\begin{table}[th!]
\caption{Hyperparameter settings used in MNIST experiments for four representative SSL. All robust SSL methods (e.g., ours (WR-SSL), DS3L and UASD) are developed based on these representative SSL.} 
\centering
\begin{tabular}{lc}
\hline
 \multicolumn{2}{c}{\textbf{Shared}} \\
\hline
Learning decayed by a factor of & 0.2\\
at training iteration & 1,000 \\
coefficient = 1 (Do not use warmup) & \\
\hline 
 \multicolumn{2}{c}{\textbf{Supervised}} \\
\hline
Initial learning rate & 0.003 \\
\hline
 \multicolumn{2}{c}{$\Pi$-\textbf{Model}} \\
\hline
Initial learning rate &0.003\\
Max consistency coefficient & 20\\
\hline
 \multicolumn{2}{c}{\textbf{Mean Teacher}} \\
 \hline
Initial learning rate & 0.0004 \\
Max consistency coefficient & 8 \\ 
Exponential moving average decay & 0.95 \\
\hline
\multicolumn{2}{c}{\textbf{VAT}} \\
\hline
Initial learning rate &0.003 \\
Max consistency coefficient &0.3 \\
VAT $\epsilon$ & 3.0 \\
VAT $\xi$ & $10^{-6}$ \\
\hline
\multicolumn{2}{c}{\textbf{Pseudo-Label}} \\
\hline
Initial learning rate& 0.0003 \\
Max consistency coefficient &1.0 \\
Pseudo-label threshold &0.95 \\
\hline
\end{tabular}
\label{hyper_mnist}
\vspace{-8mm}
\end{table}

\begin{figure*}[!t]
    \centering
    \begin{subfigure}[b]{0.2\textwidth}
        \centering
        \includegraphics[width=\linewidth]{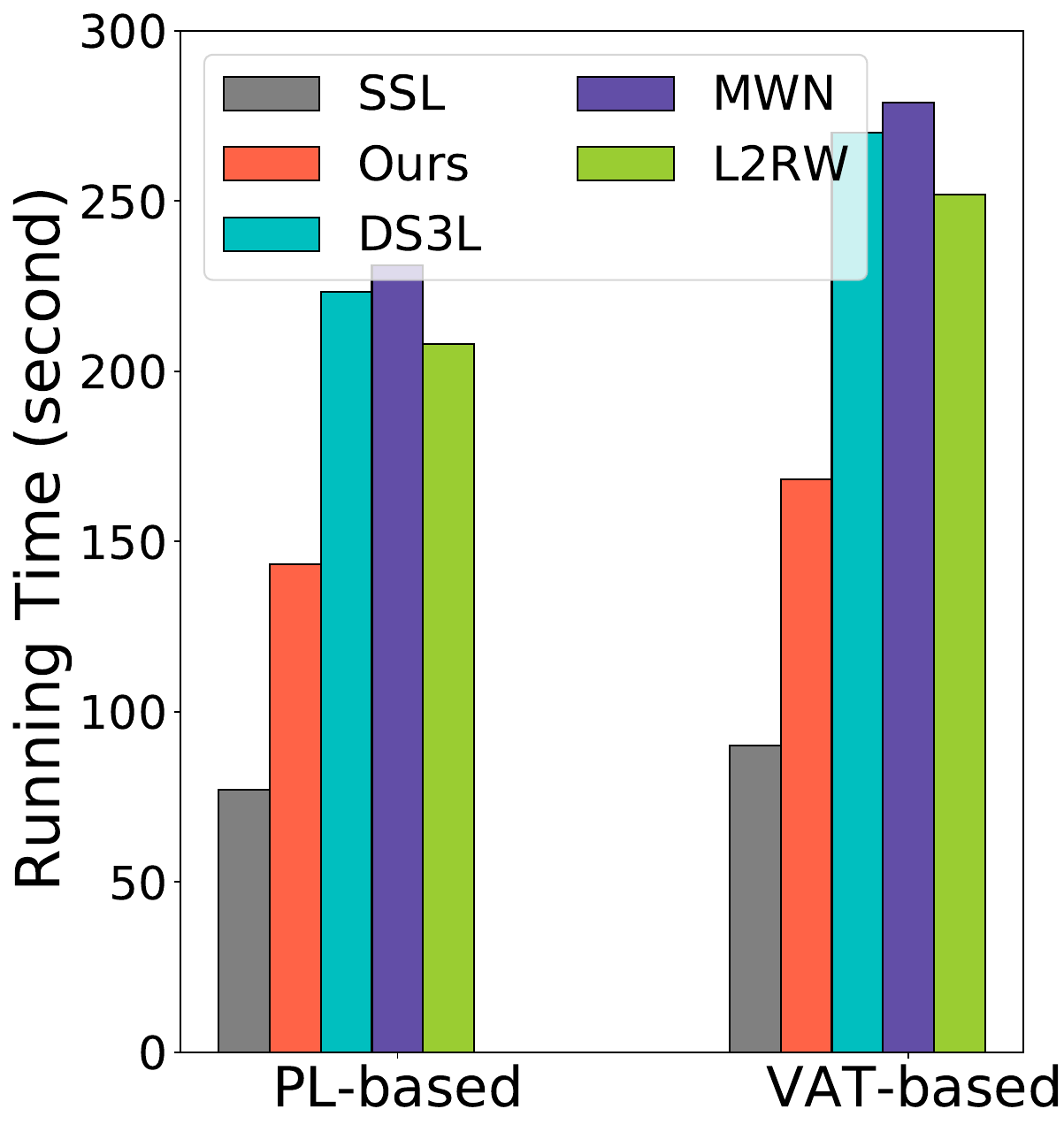}
        \caption{MNIST.}
    \end{subfigure}
    \begin{subfigure}[b]{0.2\textwidth}
        \centering
        \includegraphics[width=\linewidth]{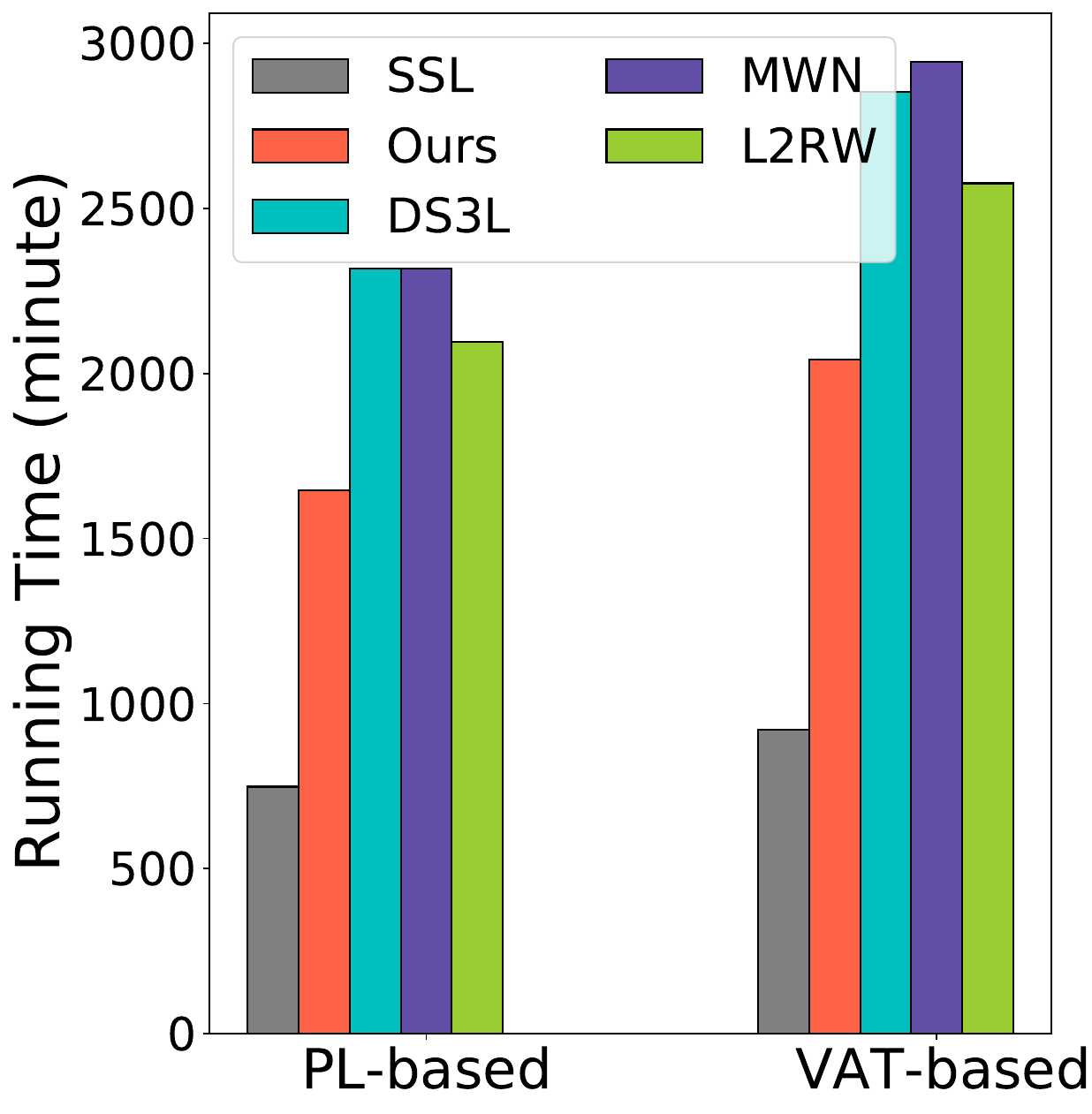}
        \caption{CIFAR10.}
    \end{subfigure}
    \begin{subfigure}[b]{0.175\textwidth}
        \centering
        \includegraphics[width=\linewidth]{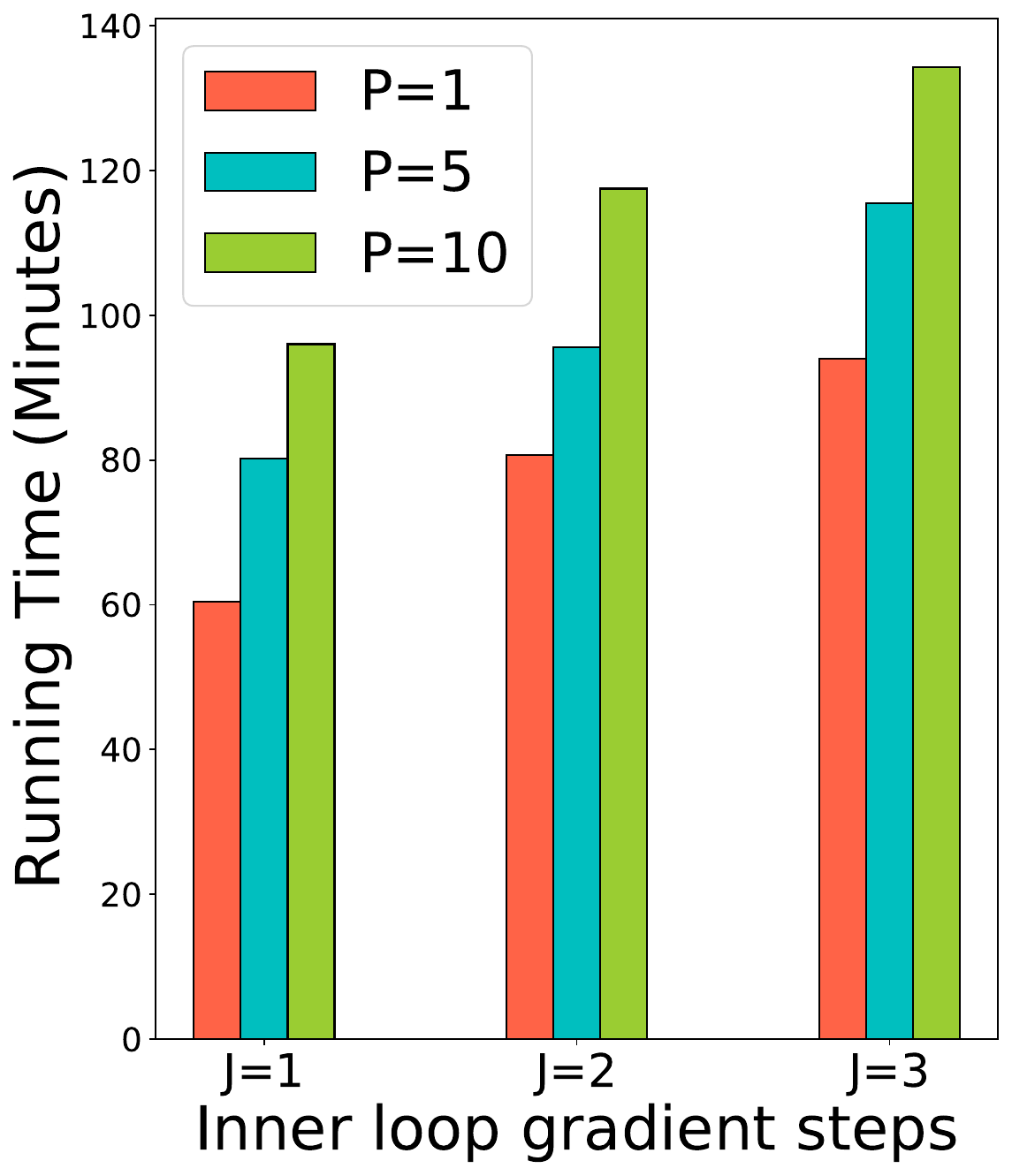}
        \caption{Running Time.}
    \end{subfigure}
    \begin{subfigure}[b]{0.175\textwidth}
        \centering
        \includegraphics[width=\linewidth]{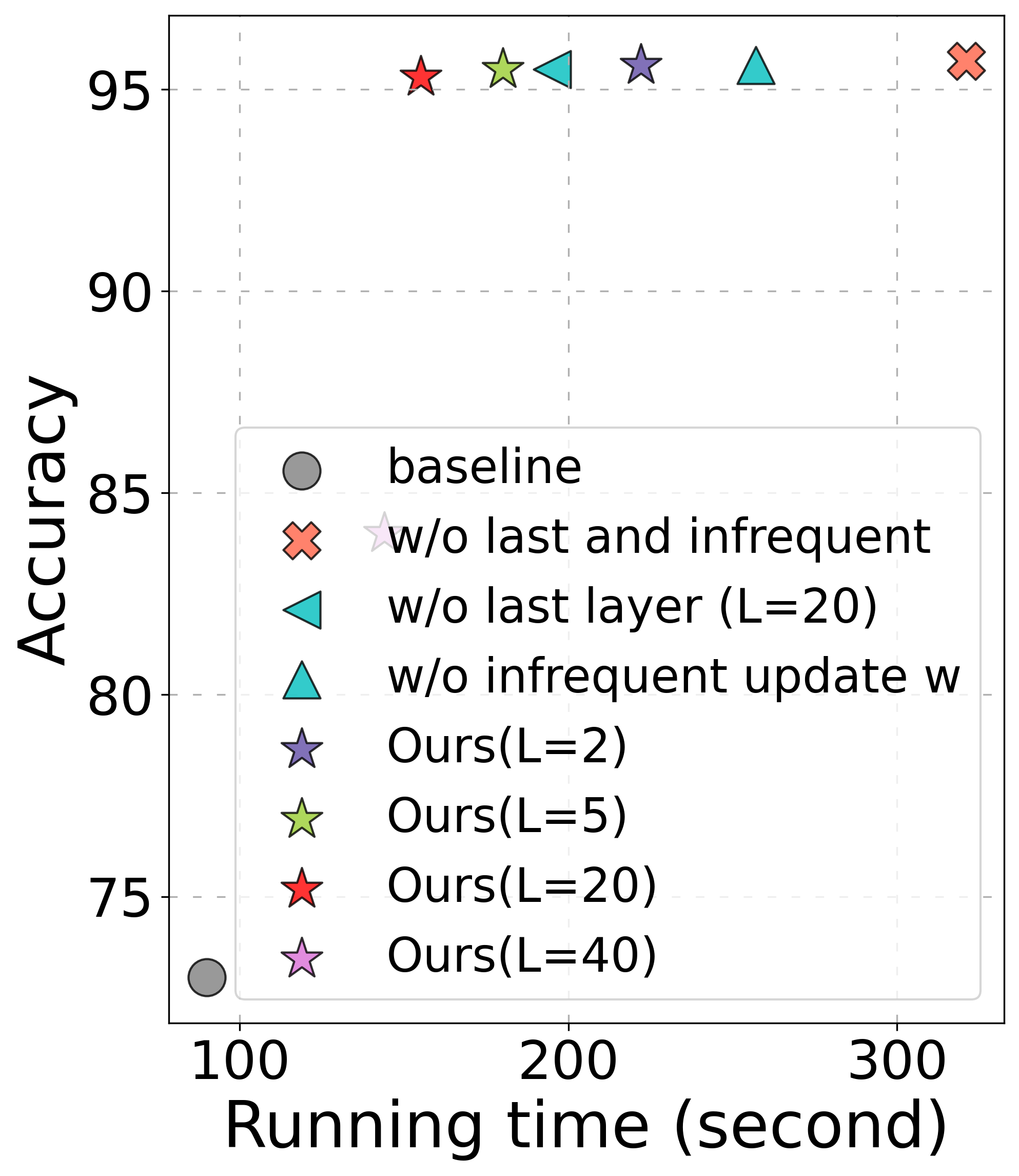}
        \caption{MNIST.}
    \end{subfigure}
    \begin{subfigure}[b]{0.175\textwidth}
        \centering
        \includegraphics[width=\linewidth]{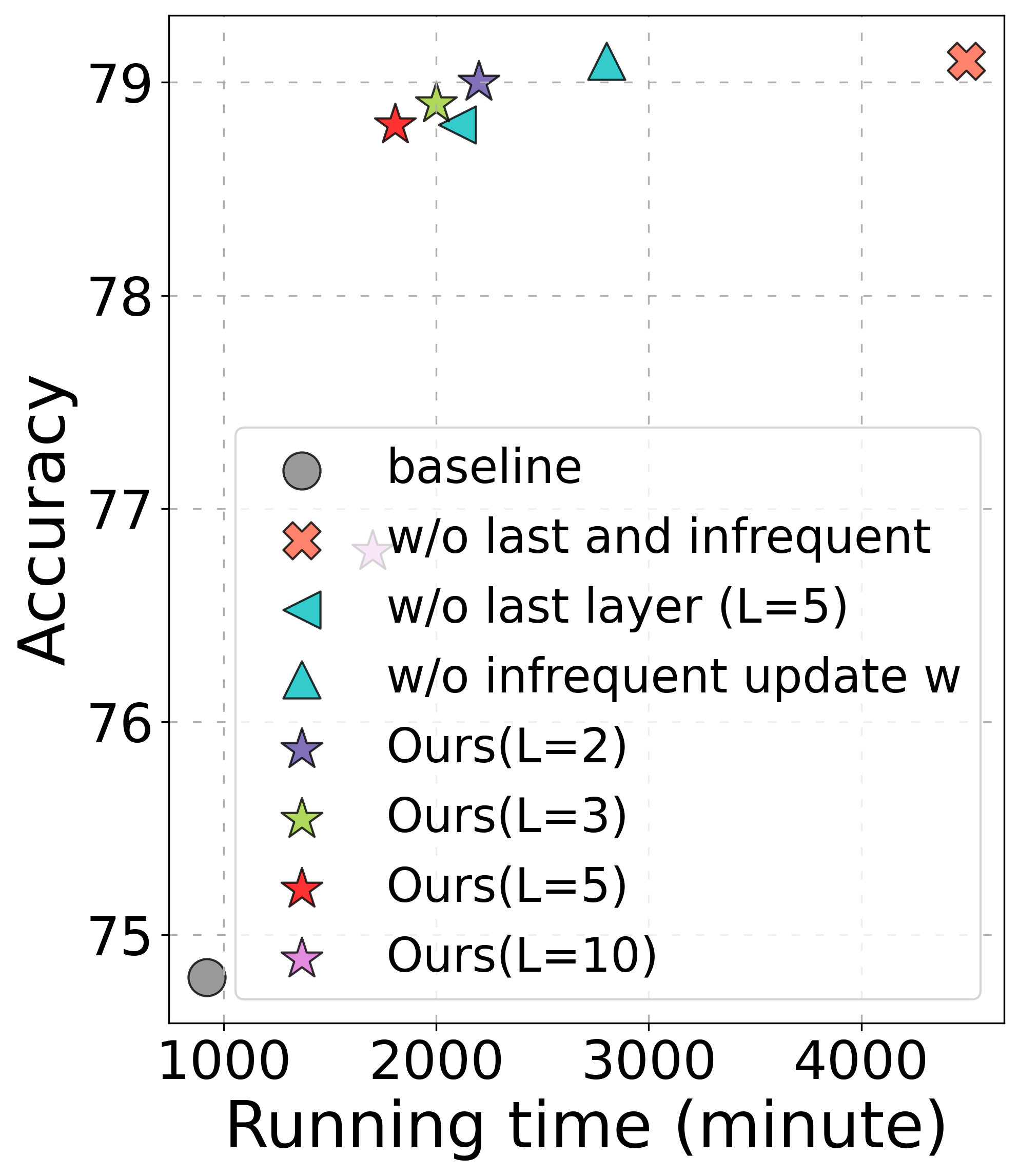}
        \caption{CIFAR10.}
    \end{subfigure}
    \vspace{-2mm}
    \small{
    \caption{Running time results. (a)-(b) show our proposed approaches are only $1.7\times$ to $1.8\times$ slower compared base SSL algorithms, while other robust SSL methods are $3\times$ slower. (c) shows that the running time of our method would increase with $J$ (inner loop gradients steps) and $P$ (inverse Hessian approximation) increase.
    (d)-(e) show running time of our strategies with different combinations of tricks viz; last layer updates and updating weights every $L$ iterations. Note that by using only last layer updates, our strategies are around $2\times$ slower. With $L = 5$ and last layer updates, we are around $1.7 \times$ to $1.8 \times$ slower with comparable test accuracy.}
    \label{fig:run time}
    }
    \vspace{-4mm}
\end{figure*}


\section{Results and Discussion}
\subsection{Performance with different OOD datasets}

In all experiments, we report the performance over five runs. Denote OOD ratio= $ \mathcal{U}_{ood}/(\mathcal{U}_{ood}+ \mathcal{U}_{in})$ where $\mathcal{U}_{in}$ is ID unlabeled set,  $\mathcal{U}_{ood}$ is OOD unlabeled set, and $\mathcal{U}=\mathcal{U}_{in} + \mathcal{U}_{ood}$. The following experimental results on each dataset are to answer all \textbf{Questions} given in Sec.~\ref{experiment}.

\begin{wrapfigure}{R}{0.21\textwidth}
  \vspace{-4mm} 
  \centering
  \includegraphics[width=0.2\textwidth]{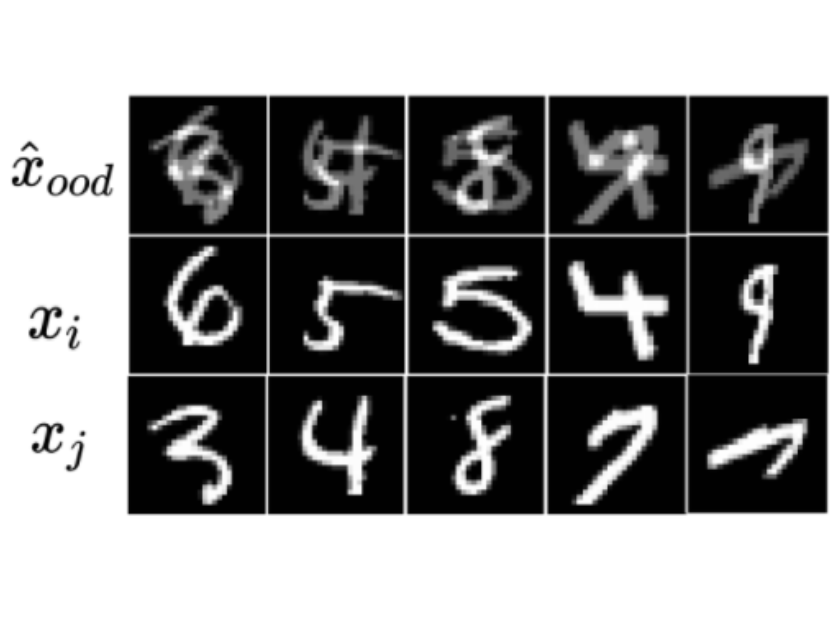}
  \vspace{-5mm}
  \caption{\footnotesize{{Examples of Mean MNIST. $\hat{x}_{ood} = (x_i+x_j)/2$} where $x_i, x_j$ are two samples from different MNIST classes.}}
  \label{fig:Mean_MNIST}
  \vspace{-3mm}
\end{wrapfigure}

In our experiments, we consider five different OOD datasets: Fashion MNIST, Mean MNIST, a subset of CIFAR-10, a subset of CIFAR-100, and a subset of SVHN. The exact portions of OODs that belong to faraway OODs and boundary OODs, respectively, can not be shown explicitly as we did not find any existing methods can estimate the information. We roughly consider Fashion MNIST as a relatively faraway OODs since grayscale clothes images in Fashion MNIST are different from the digital number in MNIST. Moreover, we consider Mean MNIST (where the instances are mean images of two MNIST classes) as relative boundary OODs because the new fused images do not belong to MNIST class but contain very similar feature vectors (see Fig~\ref{fig:Mean_MNIST}). For the subset OODs from CIFAR-10, CIFAR-100, and SVHN, we roughly consider them as relatively mixed type OODs. The subset may contain some classes close to in-distribution classes (e.g., seal v.s. whale) and some very different from in-distribution classes (e.g., flower v.s. fish).

We begin by running our algorithms on MNIST (i.e., MNIST as ID) with different types of OOD instances. First, we use  Fashion MNIST~\cite{xiao2017fashion} as OOD, which are grayscale article images. In this case, we compare all algorithms with batch normalization. As shown in Fig~\ref{fig:4} (a), the performance of the existing SSL method decreases rapidly with an increase in OOD ratio. In contrast, our approach can still maintain clear performance improvement -- e.g., our approach outperforms the base VAT SSL algorithm by almost 18\% when OOD ratio = 50\%. Compared with other robust SSL methods, our methods improve accuracy and suffer much less degradation under a high OOD ratio. Next, we use Mean MNIST (where the instances are mean images of two classes) as the OOD data.  
As shown in Fig~\ref{fig:4} (b), we see a similar pattern that the accuracy of existing SSL methods decreases when the OOD ratio increases. Across different OOD ratios, our method significantly outperforms all baselines and the base SSL methods -- e.g., 8\% increase with our approach over DS3L when OOD ratio = 50\%. 

We now study our algorithm's performance on other datasets, including CIFAR-10, CIFAR-100, and SVHN. Similar to~\cite{oliver2018realistic}, we freeze BN layers for all the methods. Recall for CIFAR-10, CIFAR-100, and SVHN datasets, we use a subset of the classes as ID and the rest of the classes (from the same dataset) as OOD. 
The average accuracy of all compared methods v.s. OOD ratio is plotted in Fig~\ref{fig:4} (c)-(d) for CIFAR-10. Our approach consistently outperforms existing baselines (and also the PL algorithm) by almost 4.5\% when the OOD ratio is 75\%. In addition, we compared our approach with the DS3L and UASD (SOTA robust SSL methods) on CIFAR100 and SVHN datasets and got a similar pattern. The results are shown in Table~\ref{SVHN} and \ref{CIFAR100}. Furthermore, in cases where we can not apply clustering (CRW), we also show the performance of our method without CRW. The result in Table~\ref{SVHN} and \ref{CIFAR100} shows that shows that our approach still performs better than DS3L and UASD, which demonstrate that our approach can achieve stable performance.

\begin{table}[h!]
\caption{SVHN-Extra with different OOD ratio.} 
\vspace{-2mm}
\centering
\begin{tabular}{c|ccc}

\textbf{OOD ratio} &   \textit{25\%} & \textit{50\%} & \textit{75\%} \\
\hline
VAT & 94.1$\pm$ 0.5 &93.6$\pm$ 0.7 &92.8$\pm$ 0.9   \\
L2RW-VAT & 96.0$\pm$ 0.6 &93.5$\pm$ 0.8 &92.7$\pm$ 0.8   \\
MWN-VAT & 96.2$\pm$ 0.5 &93.8$\pm$ 1.1 &93.0$\pm$ 1.3   \\
UASD-VAT & 96.3$\pm$ 0.5 &94.2$\pm$ 0.9 &93.3$\pm$ 1.3   \\
DS3L-VAT & 96.4$\pm$ 0.7 &93.9$\pm$ 1.0 &92.9$\pm$ 1.2   \\
Ours w/o CRW & 96.6$\pm$ 0.6 &94.4$\pm$ 0.8 &93.2$\pm$ 1.1   \\
Ours-VAT &\textbf{96.8$\pm$ 0.7} &\textbf{95.2$\pm$ 0.9} &\textbf{94.9$\pm$ 1.3} \\
\end{tabular}
\vspace{-6mm}
\label{SVHN}
\end{table}

\begin{table}[h!]
\caption{CIFAR100 with different OOD ratio.} 
\centering
\vspace{-2mm}
\begin{tabular}{c|ccc}
 
\textbf{OOD ratio} &   \textit{25\%} & \textit{50\%} & \textit{75\%} \\
                                    \hline
                 UASD-MT   &60.5$\pm$ 0.6 &60.3$\pm$ 0.7 &58.5$\pm$ 1.0   \\
                  DS3L-MT   &60.8$\pm$ 0.5 &60.1$\pm$ 1.1 &57.2$\pm$ 1.2   \\
                 Ours w/o CRW &61.5$\pm$ 0.4 &60.7$\pm$ 0.6 &59.0$\pm$ 0.8 \\
                 Ours-MT &\textbf{62.1$\pm$ 0.5} &\textbf{61.0$\pm$ 0.5} &\textbf{59.7$\pm$ 0.9} \\
                 \hline
                 UASD-PI   &61.1$\pm$ 0.5 &60.0$\pm$ 0.9 &58.4$\pm$ 1.0   \\
                  DS3L-PI   &60.5$\pm$ 0.6 &60.1$\pm$ 1.0 &57.4$\pm$ 1.3   \\
                 Ours w/o CRW &61.2$\pm$ 0.4 &60.4$\pm$ 0.4 &58.9$\pm$ 0.6 \\
                 Ours-PI &\textbf{61.6$\pm$ 0.4} &\textbf{60.7$\pm$ 0.5} &\textbf{59.5$\pm$ 0.7} \\
\end{tabular}
\vspace{-6mm}
\label{CIFAR100}
\end{table}

\begin{figure*}[!t]
    \centering
    \begin{subfigure}[b]{0.245\textwidth}
        \centering
        \includegraphics[width=\linewidth]{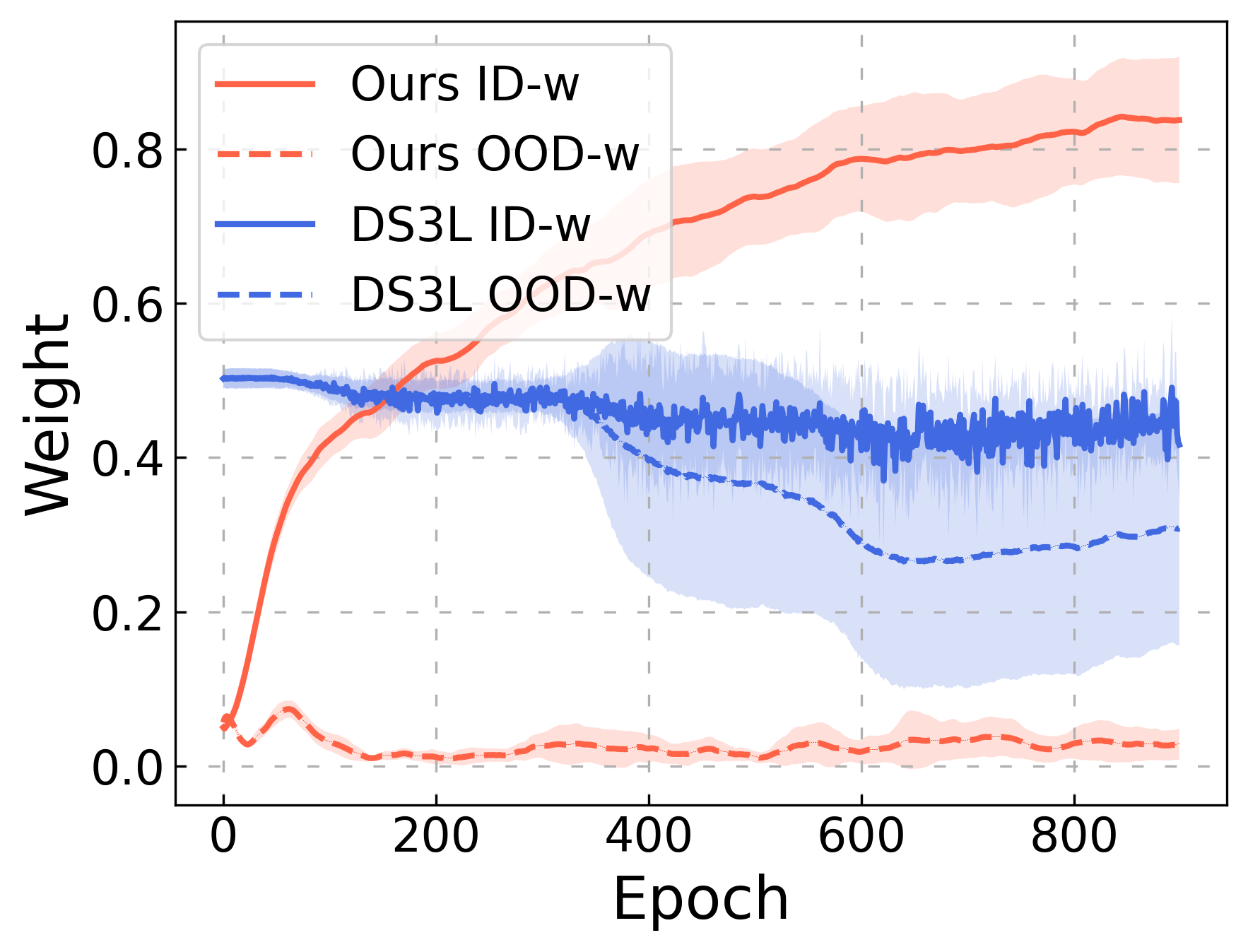}
        \caption{Weight variation curves.}
        \label{fig6a}
    \end{subfigure}
    \begin{subfigure}[b]{0.245\textwidth}
        \centering
        \includegraphics[width=\linewidth]{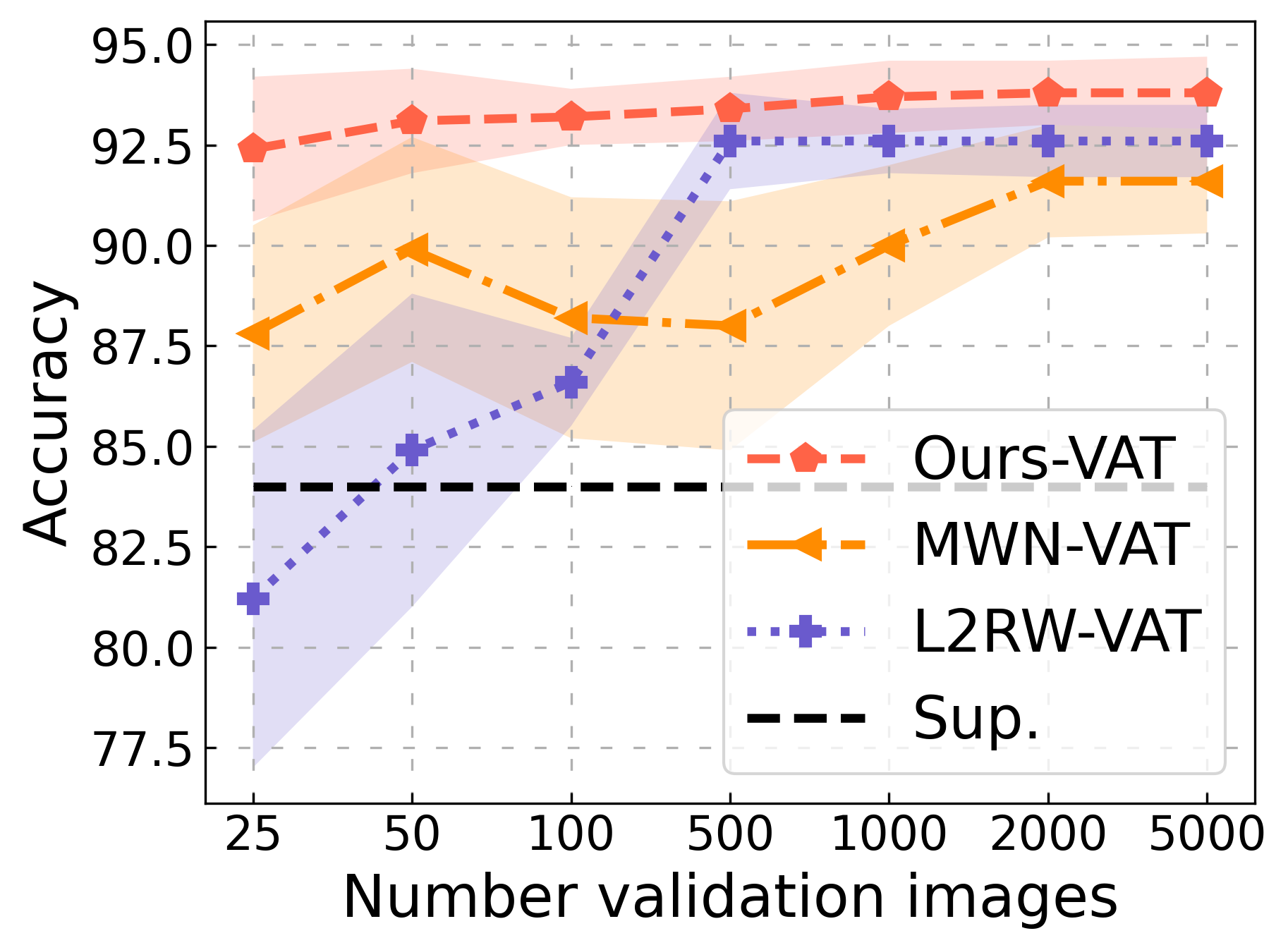}
        \caption{Effect of validation set.}
        \label{fig6b}
    \end{subfigure}
    \begin{subfigure}[b]{0.16\textwidth}
        \centering
        \includegraphics[width=\linewidth]{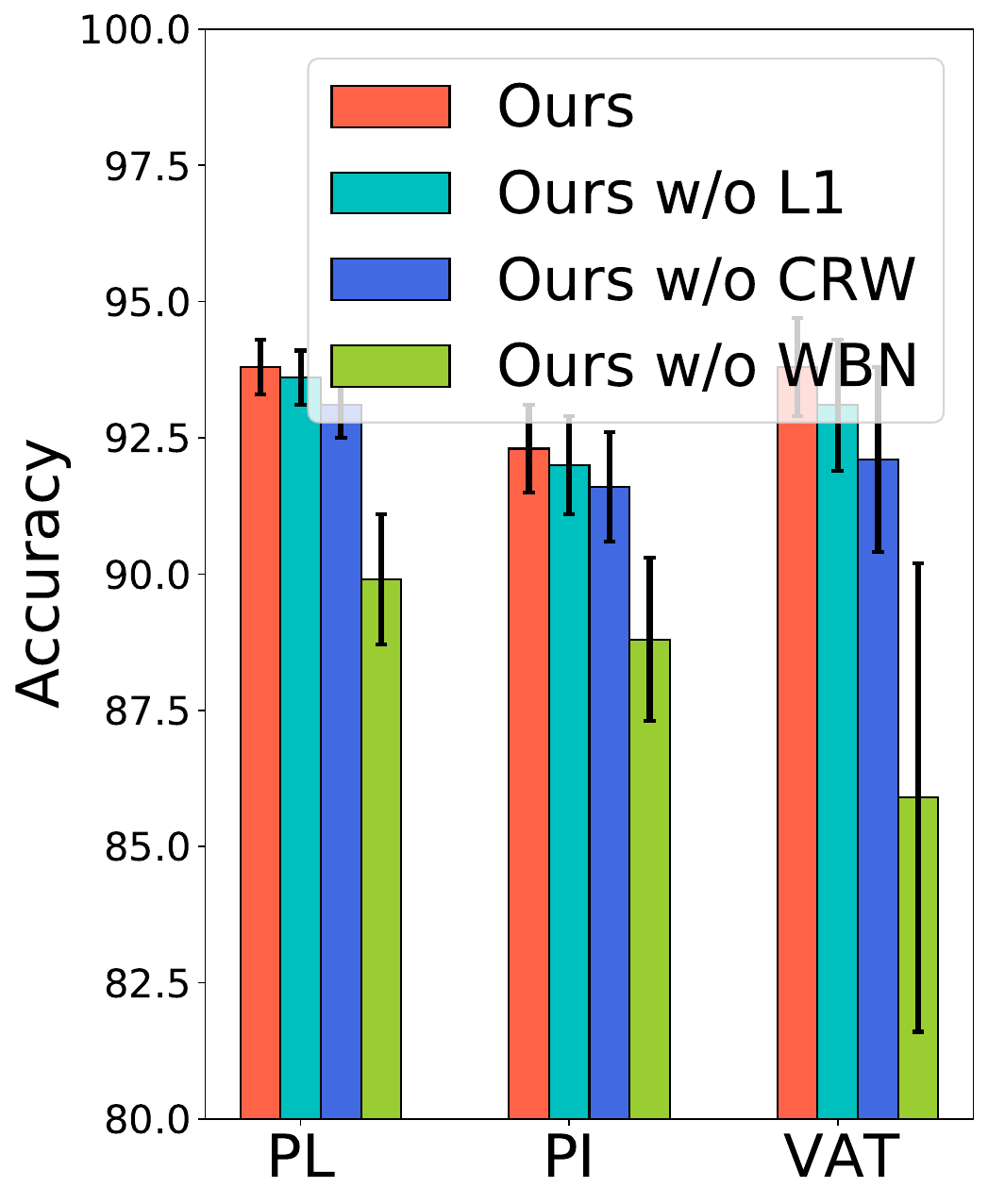}
        \caption{Ablation Study.}
        \label{fig6c}
    \end{subfigure}
    \begin{subfigure}[b]{0.15\textwidth}
        \centering
        \includegraphics[width=\linewidth]{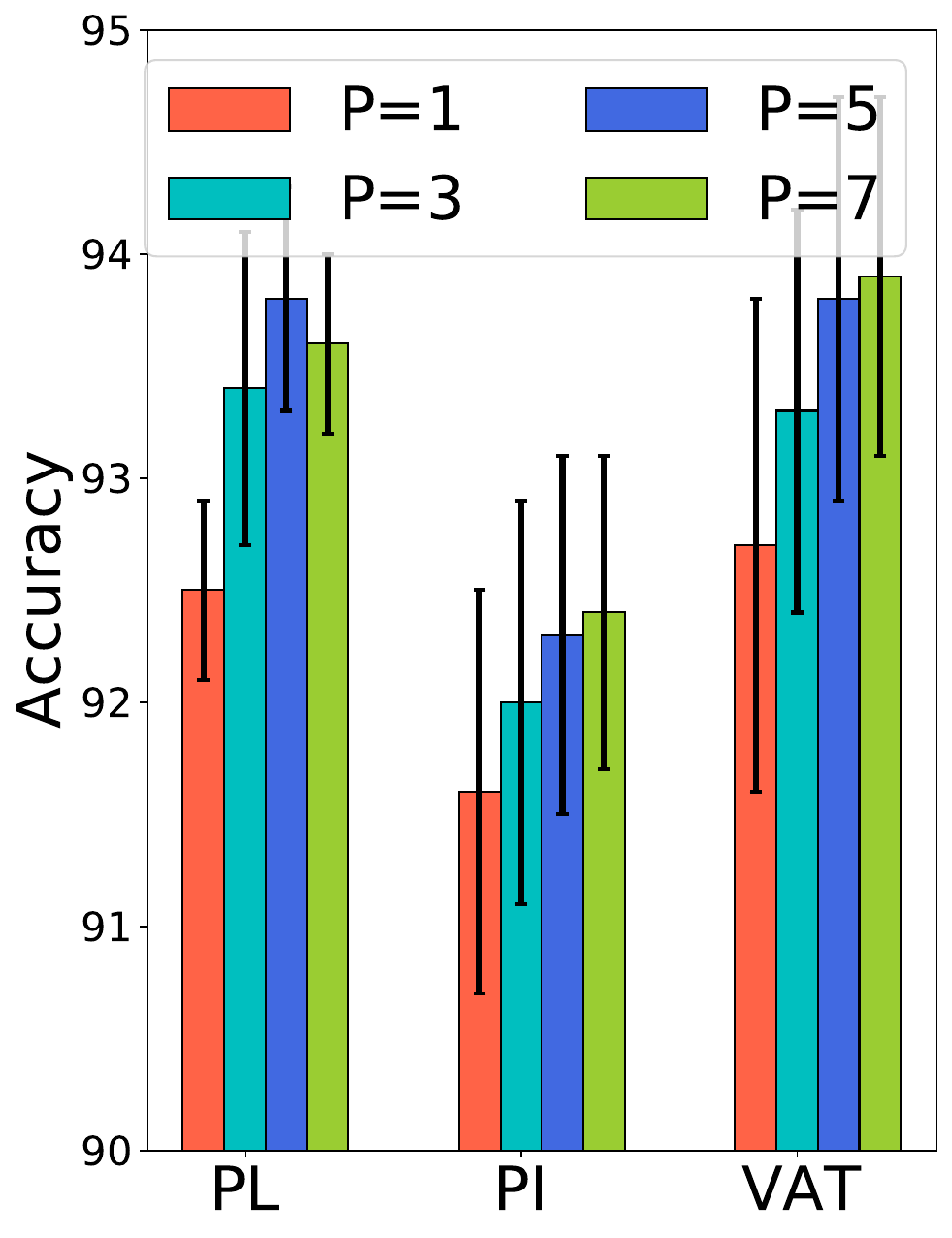}
        \caption{Ablation Study}
        \label{fig6d}
    \end{subfigure}
    \begin{subfigure}[b]{0.15\textwidth}
        \centering
        \includegraphics[width=\linewidth]{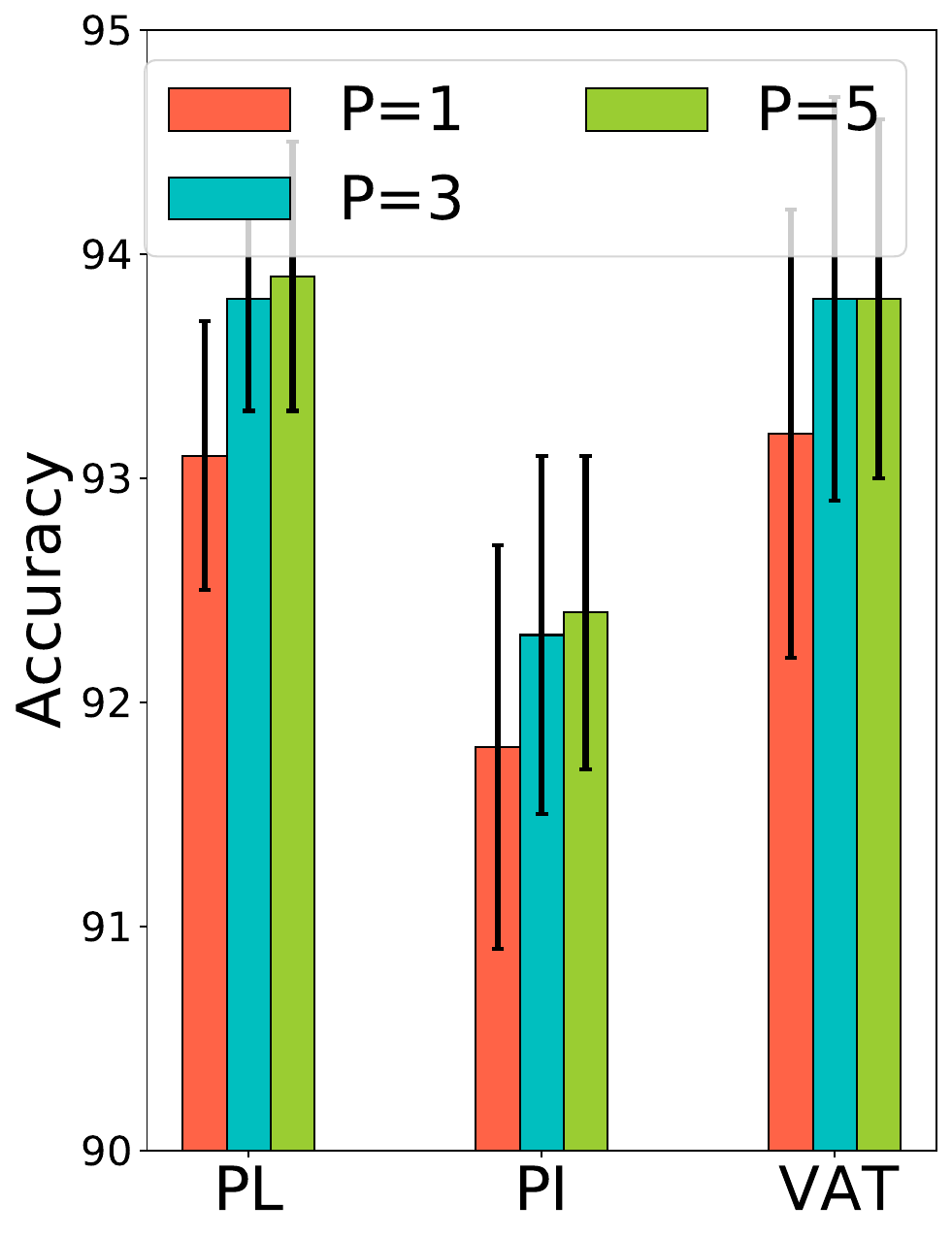}
        \caption{Ablation Study}
        \label{fig6e}
    \end{subfigure}
    \small{
    \caption{(a) shows that our method learns optimal weights for ID and OOD samples; (b) shows that our method is stable even for small validation set containing 25 images; (c) shows that WBN and CRW (or L1 regularization) are critical in retaining the performance gains of reweighting; (d)-(e) demonstrate that the performance of our approach would increase with (inner loop gradients steps) and inverse Hessian approximation) increase due to high-order approximation.}
    \label{fig:5} }
\vspace{-5mm}    
\end{figure*}

\subsection{Efficiency Analysis}
To evaluate the efficiency of our proposed approach, we first compare the running time among all methods. Fig~\ref{fig:run time} (a)-(b) shows the running time (relative to the original SSL algorithm) for MNIST and CIFAR-10. We see that our proposed approaches are only $1.7\times$ to $1.8\times$ slower than the original SSL algorithm, while other robust SSL methods (L2RW and DS3L) are almost $3\times$ slower. We note that our implementation tricks can also be applied to these other techniques (DS3L, L2RW, MWN), but this would possibly degrade performance since these approaches' performance is worse than ours even without these tricks. To further analyze our proposed speedup strategies, we plot the running time v.s. accuracy in Fig~\ref{fig:run time} (d)-(e) for different settings (with/without last and with/without infrequent updates). As expected, the results show that, without the only last layer updates and the infrequent updates (i.e. if $L = 1$), Algorithm~\ref{algorithm_robustssl} is $3 \times$ slower than SSL baseline. Whereas with the last layer updates, it is around $2\times$ slower. We get the best trade-off between speed and accuracy considering both $L = 5$ and the last layer updates. In addition, we analyze the efficiency of our approach with varying inner loop gradients steps ($J$) and inverse Hessian approximation ($P$). The result shows that the running time of our method would increase with $J$ and $P$ increase, we choose best trade-off between speed and accuracy considering $J = 3$ and $P = 5$.

\subsection{Additional Analysis}

\noindent {\bf Analysis of weight variation.} Fig~\ref{fig:5} (a) shows the weight learning curve of our approach and DS3L on Fashion-MNIST OOD. The results show that our method learns better weights for unlabeled samples compared to DS3L. The weight distribution learned in other cases are also similar. 

\noindent {\bf Size of the clean validation set.}  We explore the sensitivity of the clean validation set used in robust SSL approaches on
Mean MNIST OOD. Fig~\ref{fig:5} (b) plots the classification performance with varying the size of the clean validation set. Surprisingly, our methods are stable even when using only 25 validation images, and the overall classification performance does not grow after having more than 1000 validation images. 

\noindent {\bf Ablation Studies.} We conducted additional experiments on
Fashion-MNIST OOD (see Fig~\ref{fig:5} (c)-(e)) in order to demonstrate the contributions of the key technical components, including Cluster Re-weight (CRW) and weighted Batch Normalization (WBN). The key findings obtained from this experiment are \textit{1)} WBN plays a vital role in our weighted robust SSL framework to improve the robustness of BN against OOD data; \textit{2)} Removing CRW (or L1 regularization) results in performance decrease, especially for VAT based approach, which demonstrates that CRW (and L1 regularization) can further improve performance for our robust-SSL approach;
\textit{3)} Fig~\ref{fig:5} (d)-(e) demonstrate that the performance of our approach would increase with (inner loop gradients steps) and inverse Hessian approximation) increase due to high-order approximation. Based on this, we choose best trade-off between speed and accuracy considering $J=3$ and $P=5$ when considering running time analysis in Fig~\ref{fig:run time} (c).

\noindent {\bf L1 vs CRW tricks. } Next, we discuss the trade-offs between L1 and CRW. 
We first analyzed the sensitivity of our proposed CRW methods to the number of clusters used. Table~\ref{OOD_ex:K} demonstrates the test accuracies of our approach with varying numbers of clusters. The results indicate a low sensitivity of our proposed methods to the number of clusters. We also find that CRW generally can further improve the performance. Part of this success can be attributed to the good pretrained features. Our approach without CRW (only consider L1 regularization to reduce overfitting) performs well even without this additional information ( Table~\ref{SVHN} ,\ref{CIFAR100}, and Fig.~\ref{fig:5} (c)). 

\begin{table}[th!]
\vspace{-3mm}
\caption{Test accuracies for different numbers of clusters $K$ on the MNIST dataset with 50\%  Mean MNIST as OODs .} 
\centering
\begin{tabular}{c|cccc}
\# Clusters & K=5 & K=10& K=20 & K =30 \\
\hline
Ours-VAT &94.7$\pm$ 0.7 &95.3$\pm$ 0.4 &96.3$\pm$ 0.5  &95.6$\pm$ 0.5 \\
\hline
Ours-PL &95.2$\pm$ 0.5 &95.3$\pm$ 0.5 &96.2$\pm$ 0.4  &95.9$\pm$ 0.5 
\end{tabular}
\label{OOD_ex:K}
\end{table}

\vspace{-6mm}

\section{Conclusion}
In this work, we first propose the research question: \textit{How out-of-distribution data hurt semi-supervised learning performance?} To answer this question, 
we study the impact of OOD data on SSL algorithms and demonstrate empirically that the SSL algorithms' performance depends on how close the OOD instances are to the decision boundary (and the ID data instances). To address the above causes, we proposed a novel unified weighted robust SSL framework, which is designed to improve the robustness of BN against OODs. To address the limitation of low-order approximations in bi-level optimization (DS3L), we designed an implicit-differentiation based algorithm that considered high-order approximations of the objective and is scalable to a higher number of inner optimization steps to learn a massive amount of weight parameters. 
In addition, we conduct a theoretical analysis about the impact of faraway OODs in the BN step and discuss the connection between our approach (high-order approximation based on implicit differentiation) and low-order approximation approaches.
We show that our weighted robust SSL approach significantly outperforms existing robust approaches (L2RW, MWN, Safe-SSL, and UASD)  on several real-world datasets.

\vspace{-2mm}
\section*{Acknowledgments}
\vspace{-1mm}

The work of Xujiang Zhao and Feng Chen was supported by the National Science Foundation (NSF) under grant numbers 2147375, 2107449, and 1954376.
Rishabh Iyer and Killamsetty Krishnateja would like to acknowledge support from NSF Grant Number IIS-2106937, a gift from Google Research, and the Adobe Data Science Research award.

\vspace{-2mm}

\small{
\bibliographystyle{IEEETranSN}
\bibliography{references}
}


\end{document}